%% file: main.tex
\documentclass{article}

\usepackage{color}
\usepackage[table]{xcolor}
\definecolor{myredd}{RGB}{176,35,24}
\definecolor{mygreenn}{RGB}{78,173,91}
\definecolor{mybluee}{RGB}{46,112,186}
\definecolor{myred}{RGB}{255,225,220}
\definecolor{mygreen}{RGB}{194,244,210} 
\definecolor{myblue}{RGB}{216,236,255}
\definecolor{byzantine}{rgb}{0.74, 0.2, 0.64}
\definecolor{table-blue}{RGB}{173, 216, 230}

\newcommand{\highlightbox}[1]{
\begin{tcolorbox}[
  colback=myblue!50!white,  %
  boxrule=0pt,          %
  frame empty,
  boxrule=0pt,
  colframe=white,
  width=\textwidth,     %
  left=3pt,             %
  right=0pt             %
]
#1
\end{tcolorbox}
}

\PassOptionsToPackage{numbers, compress}{natbib}

\usepackage[preprint]{neurips_2025}

\usepackage[utf8]{inputenc} %
\usepackage[T1]{fontenc}    %
\usepackage{hyperref}       %
\usepackage{url}            %
\usepackage{booktabs}       %
\usepackage{amsfonts}       %
\usepackage{nicefrac}       %
\usepackage{microtype}      %
\usepackage{xcolor}         %
\usepackage{bbm}
\usepackage{soul}
\usepackage{multicol} %
\usepackage[font=small]{caption}
\usepackage{titletoc}

\newcommand{\1}{\mathbb{1}}

\newcommand{\pitheta}{\pi_\theta}
\newcommand{\var}{\text{Var}}

\input{math_commands}

\usepackage{dsfont}
\usepackage{amssymb}
\usepackage{amsthm}
\usepackage{wrapfig}

\usepackage[most,skins,theorems]{tcolorbox}
\tcbset{
  aibox/.style={
    width=\linewidth,
    top=7pt,
    bottom=2pt,
    colback=blue!6!white,
    colframe=black,
    colbacktitle=black,
    enhanced,
    center,
    attach boxed title to top left={yshift=-0.1in,xshift=0.15in},
    boxed title style={boxrule=0pt,colframe=white,},
  }
}
\newtcolorbox{AIbox}[2][]{aibox,title=#2,#1}

\definecolor{rliableolive}{HTML}{BBCC33}
\definecolor{rliableblue}{HTML}{77AADD}
\definecolor{rliablered}{HTML}{EE8866}

\theoremstyle{definition}

\newtheorem{template}{Template}

\usepackage{cleveref}
\crefname{section}{Sec.}{Sec.}
\crefname{theorem}{Theorem}{Theorems}
\crefname{corollary}{Corollary}{Corollaries}
\crefname{lemma}{Lemma}{Lemmas}
\crefname{equation}{Eq.}{Eq.}
\crefname{proposition}{Proposition}{Propositions}
\crefname{claim}{Claim}{Claims}
\crefname{remark}{Remark}{Remarks}
\crefname{observation}{Observation}{Observations}
\crefname{assumption}{Assumption}{Assumptions}
\crefname{template}{Template}{Template}
\crefname{definition}{Definition}{Definitions}
\crefname{appendix}{Appendix}{Appendices}
\crefname{algorithm}{Algorithm}{Algorithms}
\crefname{figure}{Fig.}{Fig.}
\crefname{table}{Table}{Tables}
\crefname{property}{Property}{Properties}
\crefname{line}{Line}{Lines}
\newtheorem{theorem}{Theorem}

\definecolor{table-blue}{RGB}{173, 216, 230}
\definecolor{darkblue}{rgb}{0, 0, 0.5}
\hypersetup{colorlinks=true, citecolor=darkblue, linkcolor=darkblue, urlcolor=magenta}

\usepackage{graphicx}
\usepackage{multirow}

\usepackage{xspace}
\newcommand{\method}{\textrm{VeriFree}\xspace}

\title{Reinforcing General Reasoning without Verifiers}

\renewcommand\footnotemark{}
\author{%
  Xiangxin Zhou$^{*23}$, Zichen Liu$^{*14}$, Anya Sims$^{*15}$, Haonan Wang$^{14}$, Tianyu Pang$^{1}$, \\
  \textbf{Chongxuan Li$^{6}$, Liang Wang$^{23}$, Min Lin$^{1}$, Chao Du$^{\dagger1}$}\\
  \thanks{$^*$Equal contribution. Work done during Anya’s associate membership and Haonan's internship at Sea AI Lab.\looseness=-1}
  \thanks{$^\dagger$Correspondence to Chao Du.}
  $^{1}$Sea AI Lab, Singapore;
  $^{2}$University of Chinese Academy of Sciences;\\
  $^{3}$Institute of Automation, Chinese Academy of Sciences;\\
  $^{4}$National University of Singapore;
  $^{5}$University of Oxford;
  $^{6}$Renmin University of China\\
  \texttt{zhouxiangxin1998@gmail.com;}
  \texttt{\{liuzc, tianyupang, linmin, duchao\}@sea.com}
}

\begin{document}

\maketitle

\vspace{-3mm}
\input{sections/00-abstract}

\input{sections/10-intro}

\input{sections/20-method}

\input{sections/30-experiment}

\input{sections/40-related}

\input{sections/50-conclusion}

\bibliography{main}
\bibliographystyle{plainnat}

\newpage
\appendix

\input{sections/appendix/A00-related}

\input{sections/appendix/A13-proofs}
\input{sections/appendix/A20-dataset}

\input{sections/appendix/A30-extended-results}
\input{sections/appendix/A40-case-study}
\section{Broader Impact}
\label{sec:broader_impact}
Improving the general reasoning capabilities of large language models (LLMs) has the potential to significantly expand their applicability across domains such as education, scientific discovery, legal analysis, and decision support. By removing reliance on explicit verifiers—often expensive or domain-specific—our approach democratizes access to advanced reasoning systems, enabling broader use in resource-constrained settings.

However, improved reasoning can amplify both beneficial and harmful behaviors, depending on the intent and oversight of the user. As such, responsible usage, transparent reporting of model limitations, and value alignment remain critical.

We encourage future work that complements our method with safeguards for misuse prevention and performance guarantees under distributional shifts, particularly as general-purpose reasoning models become more widely integrated into real-world workflows.

\end{document}

%% file: math_commands.tex
\usepackage{amsmath,amsfonts,bm}

\def\eqref#1{equation~\ref{#1}}

\def\1{\bm{1}}

\def\vo{{\bm{o}}}

\def\vx{{\bm{x}}}
\def\vy{{\bm{y}}}
\def\vz{{\bm{z}}}

\DeclareMathAlphabet{\mathsfit}{\encodingdefault}{\sfdefault}{m}{sl}
\SetMathAlphabet{\mathsfit}{bold}{\encodingdefault}{\sfdefault}{bx}{n}

\newcommand{\E}{\mathbb{E}}

%% file: sections/00-abstract.tex
\begin{abstract}
The recent paradigm shift towards training large language models (LLMs) using DeepSeek-R1-Zero-style reinforcement learning (RL) on verifiable rewards has led to impressive advancements in code and mathematical reasoning. However, this methodology is limited to tasks where rule-based answer verification is possible and does not naturally extend to real-world domains such as chemistry, healthcare, engineering, law, biology, business, and economics.
Current practical workarounds use an additional LLM as a model-based verifier; however, this introduces issues such as reliance on a strong verifier LLM, susceptibility to reward hacking, and the practical burden of maintaining the verifier model in memory during training.
To address this and extend DeepSeek-R1-Zero-style training to general reasoning domains, we propose
a verifier-free method (\textbf{\method}) that bypasses answer verification and instead uses RL to directly maximize the probability of generating the reference answer.
We compare \method with verifier-based methods and demonstrate that, in addition to its significant practical benefits and reduced compute requirements, \method matches and even surpasses verifier-based methods on extensive evaluations across MMLU-Pro, GPQA, SuperGPQA, and math-related benchmarks. 
Moreover, we provide insights into this method from multiple perspectives: 
as an elegant integration of training both the policy and implicit verifier in a unified model, and as a variational optimization approach. Code is available at \url{https://github.com/sail-sg/VeriFree}.
\end{abstract}

%% file: sections/10-intro.tex
\begin{figure}[h]
    \vspace{-.4cm}
    \centering
    \includegraphics[width=1\linewidth]{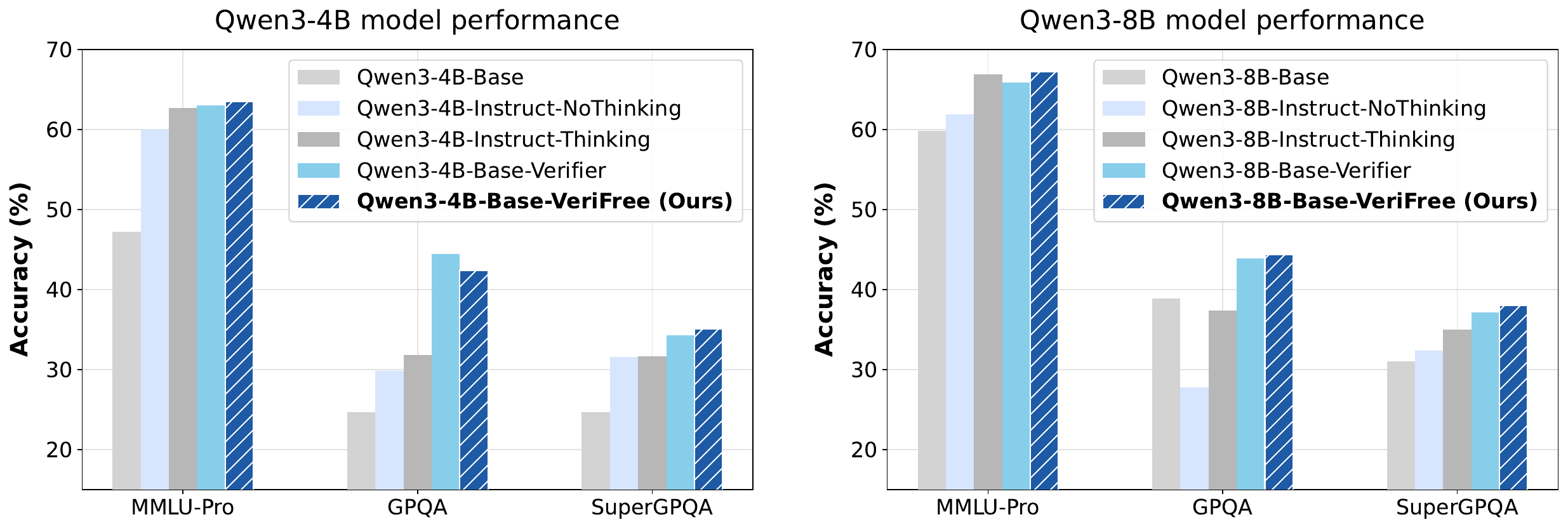}
    \caption{The general reasoning capability is significantly improved when we apply \method to fine-tune Qwen3 base models on a general reasoning dataset. Notably, \method can perform on par with or even surpass the instruct models and models RL-tuned from base with a specialized LLM as a verifier.}
    \label{fig:fig1}
\end{figure}
\section{Introduction}
\vspace{-.1cm}

DeepSeek-R1-Zero~\citep{guo2025deepseek} recently demonstrated that training large language models (LLMs) using reinforcement learning (RL) with verifiable rewards can be extremely effective in improving reasoning capabilities. In this RL with verifiable rewards (RLVR) framework~\citep{lambert2024tulu3}, the LLM generates a reasoning trace (i.e., chain of thoughts, CoT) followed by a final answer. A rule-based program then extracts and evaluates the final answer, assigning a reward of 1 to the response if the final answer is correct and 0 otherwise. The model is trained with RL using GRPO~\citep{shao2024deepseekmath}---a simplified variant of PPO~\citep{schulman2017proximal}.
\looseness=-1

The simplicity of this approach, coupled with impressive performance improvements in mathematical reasoning tasks, has sparked a wave of follow-up works in this paradigm of \textit{RL with rule-based verifiable rewards}~\citep{liu2025understanding,deepscaler2025,yu2025dapo}, which we will refer to as the \emph{R1-Zero-style training} in the following. However, these methods remain limited to domains such as mathematics and code, where rule-based verification is feasible. Reasoning is critical far beyond math and coding; however, the difficulty of answer verification in general reasoning tasks poses a major obstacle to applying this training paradigm to broader domains. To address this limitation, we investigate how to extend R1-Zero-style training to tasks where rule-based answer verification is not possible.

A natural extension, as explored in recent general reasoning works~\citep{su2025expanding,generalreasoner}, is to introduce a specialized LLM as a verifier, similar to the reward model used in RL from human feedback (RLHF)~\citep{rlhf,ouyang2022training}. In these methods, the \textit{model-based verifier} is queried to determine whether the generated answer is equivalent to the reference answer. Although this approach bypasses the need for rule-based evaluation, it introduces several potential drawbacks (as in standard RLHF): it depends on the availability of a strong verifier LLM, it converts the R1-Zero-style paradigm into optimizing a model-based reward, which makes it vulnerable to reward hacking~\citep{gao2023scaling}, and it adds significant computational overhead by requiring an additional model to be held in memory and queried during training.
\looseness=-1

\begin{figure}[t]
\centering
\includegraphics[width=0.99\columnwidth]{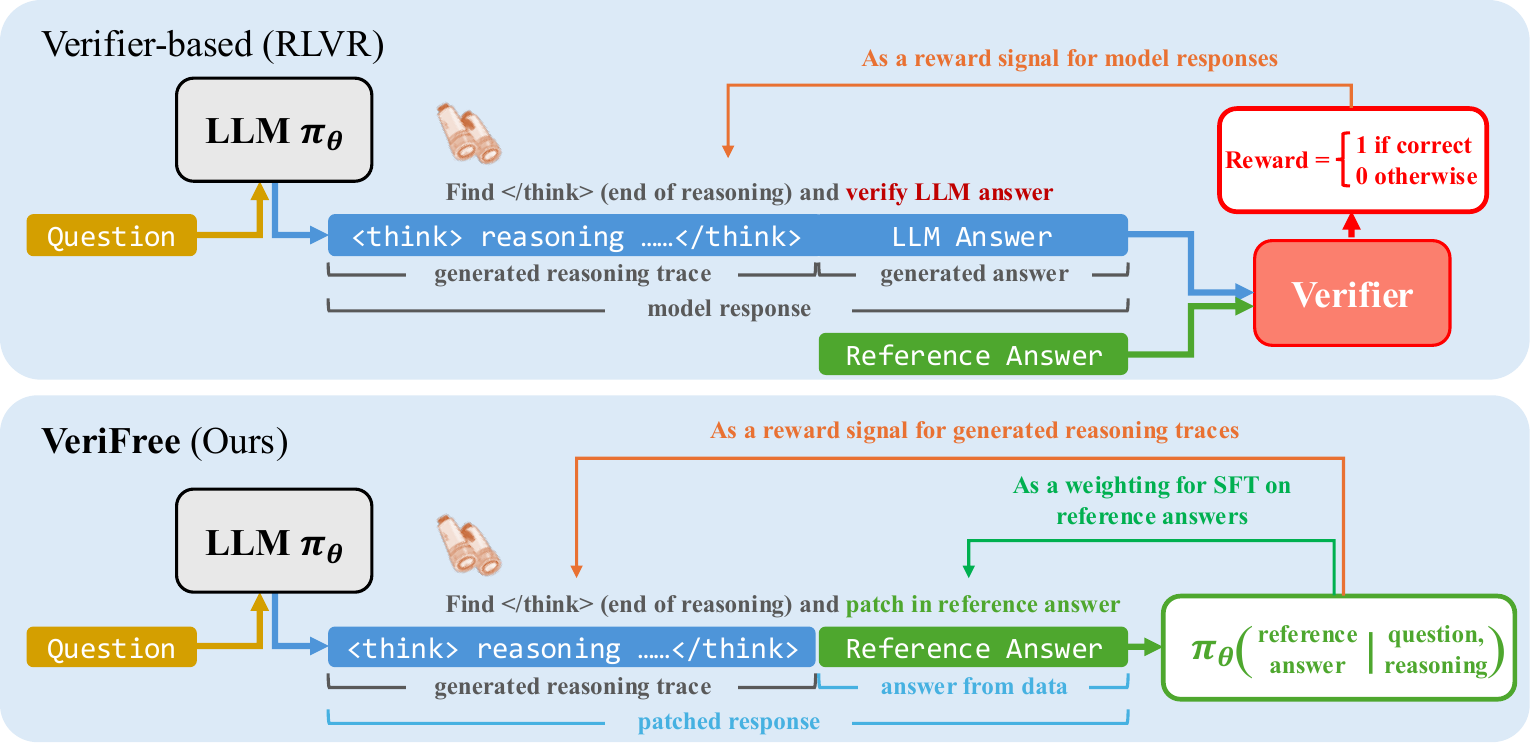}
\caption{\method enables R1-Zero-style LLM training without requiring access to a verifier. In the case of a single correct answer format, \method optimizes exactly the same objective as R1-Zero with a lower variance gradient estimator.}\label{fig:framework}
\vspace{-.3cm}
\end{figure}

In this work, we propose an alternative: a verifier-free approach that preserves the benefits of the RL paradigm while eliminating the reliance on explicit verification, either performed by rules or by models.
Our method proceeds as follows. Given a question, we only generate the reasoning trace and concatenate it with the reference answer from the dataset. We then evaluate the likelihood of the reference answer conditioned on the question and the generated reasoning trace. This likelihood serves both as a reward signal for policy gradients on the reasoning trace and as a weighting term for supervised training of the reference answer.
We term our method \method since it does not rely on rule- or model-based verifiers, and give an illustration in \Cref{fig:framework}.

This approach has several appealing properties. First, when there is a unique correct answer string, our method is equivalent in expectation to the objective in RLVR, but with lower variance, which can be viewed as a form of reward shaping~\citep{rewardshaping2,rewardshaping}. Even when multiple valid answers exist, we show empirically that using just one as a reference provides a sufficient learning signal to elicit strong reasoning behavior. Additionally, this framework can be viewed through a variational lens as a neat way of optimizing over latent reasoning traces.

To make this work in practice, we identify and address several subtle challenges, including effective variance reduction and precise handling of tokenization at the reasoning-answer patching point. We conduct comprehensive ablations to understand the impact of each design choice. We benchmark our method across a diverse set of general reasoning tasks, and the results are striking: as shown in \Cref{fig:fig1}, \method not only matches but often outperforms verifier-based alternatives, while being simpler, faster, less memory-intensive, and more robust.

%% file: sections/20-method.tex
\vspace{-.1cm}
\section{Methodology}
\vspace{-.1cm}
\subsection{Preliminaries: Verifier-Based Reinforcement Learning}
\vspace{-.1cm}
\label{subsec:method}
In RL applied to LLMs, the language model is treated as a policy $\pi_\theta$ that generates an output $\vo$ autoregressively in response to an input question $\vx$. The goal is typically to optimize $\pi_\theta$ to maximize a given reward function $R(\vx, \vo)$:
\begin{equation}
    \label{eq:rl_objective}
    \theta \in \underset{\theta}{\arg\max} \, \E_{\vo \sim \pi_\theta(\cdot|\vx)}\left[ R(\vx, \vo) \right].
\end{equation}
In R1-Zero-style RL, the reward is computed by first parsing the response $\vo$ into a reasoning trace $\vz$ and a final answer $\vy$. A verifier then checks $\vy$ against the ground-truth reference answer $\vy^\star$ and assigns a binary reward based on correctness, namely $R_{\text{Verifier}}(\vy; \vy^\star) = \mathds{1}_{\{\vy \equiv \vy^\star\}}$.\footnote{We use `$\equiv$' to denote semantic equivalence, where multiple expressions can be judged correct. For example, if $\vy^\star=$ `\texttt{8/5}', then `\texttt{8/5}', `\texttt{1.6}', and `\texttt{\textbackslash frac\{8\}\{5\}}' are all considered correct.}
Decomposing the model output as $\vo = (\vz, \vy)$, we can rewrite the objective in \Cref{eq:rl_objective} as:
\begin{align}\label{eq:verifier_objective}
    J_{\text{Verifier}}(\theta;\vx,\vy^\star) = \E_{\vz\sim\pitheta(\cdot|\vx)}{\E_{\vy\sim\pitheta(\cdot|\vx,\vz)}[R_{\text{Verifier}}(\vy;\vy^\star)]},
\end{align}
which separates the sampling of the reasoning trace and the final answer. To maximize this objective, the model $\pitheta$ is typically updated using the policy gradient estimator~\citep{sutton2018rlbook}:
\begin{equation}
\small
    \nabla_\theta J_{\text{Verifier}}(\theta;\vx,\vy^\star) = \underset{{\vz\sim\pitheta(\cdot|\vx)}}{\E}~\underset{\vy\sim\pitheta(\cdot|\vx,\vz)}{\E}\Big[R_{\text{Verifier}}(\vy;\vy^\star)\big[\nabla_\theta\log\pitheta(\vz|\vx) + \nabla_\theta\log\pitheta(\vy|\vx,\vz)\big]\Big].
    \label{eq:grad_verifier}
\end{equation}
However, this approach requires evaluating answer correctness via
$R_{\text{Verifier}}(\vy;\vy^\star) = \mathds{1}_{\{\vy \equiv \vy^\star\}}$, which is often nontrivial.
While in domains such as math and code, this evaluation is feasible via rules~\citep{guo2025deepseek,liu2025understanding} or test cases~\citep{gehring2024rlef}, accurate verification in general reasoning tasks is substantially more difficult.
As a result, recent advances in R1-Zero-style training have largely been restricted to verifiable domains, leaving reasoning tasks in general domains underexplored.
In light of this, we present an alternative verifier-free approach which naturally extends this training paradigm to broader reasoning domains.
\looseness=-1

\vspace{-.1cm}
\subsection{\method Policy Optimization}
\vspace{-.1cm}
We begin with the standard objective in \Cref{eq:verifier_objective} and show that, in the case of a single correct answer, we can derive an equivalent objective that does not require a verifier. Moreover, we compare the gradient estimators of this new objective and the verifier-based objective (\Cref{eq:grad_verifier}), and demonstrate that our verifier-free gradient estimator has the additional benefit of lower variance.

Starting from \Cref{eq:verifier_objective} and assuming a unique correct answer such that $R_{\text{Verifier}}(\vy;\vy^*) = \mathds{1}_{\{\vy = \vy^\star\}}$ (i.e., exact match rather than semantic equivalence $\mathds{1}_{\{\vy \equiv \vy^\star\}}$), the \method objective is derived as follows:
\begin{align}
    J_{\text{Verifier}}(\theta;\vx,\vy^\star)
    &= \E_{\vz\sim\pitheta(\cdot|\vx)}\Big[{\E_{\vy\sim\pitheta(\cdot|\vx,\vz)}[\,\overbrace{\mathds{1}_{\{\vy = \vy^\star\}}}^{R_{\text{Verifier}}}\,]}\Big]\nonumber\\
    &= \E_{\vz\sim\pitheta(\cdot|\vx)}\Big[\sum_{\vy}\pitheta(\vy|\vx,\vz)\mathds{1}_{\{\vy = \vy^\star\}}\Big]\nonumber\\
    &= \E_{\vz\sim\pitheta(\cdot|\vx)}\Big[{\qquad\:\,\underbrace{\pitheta(\vy^\star|\vx,\vz)}_{R_{\text{\method}}}\qquad\:\,}\Big]  \triangleq
    J_{\text{\method}}(\theta;\vx,\vy^\star).
    \label{eq:verifree_objective}
\end{align}
This can be interpreted as follows: if only one answer $\vy^\star$ is correct and receives a reward of 1 (while all others receive 0), then the expected reward given a reasoning trace $\vz$ can be computed directly as the probability assigned to $\vy^\star$, effectively marginalizing out $\vy$.
The corresponding gradient estimator is given by (see \Cref{proof:gradients} for a full derivation):
\begin{equation}\label{eq:grad_ours}
\footnotesize
    \nabla_\theta J_{\text{\method}}(\theta;\vx,\vy^\star) = \E_{\vz\sim\pitheta(\cdot|\vx)}\bigg[R_{\text{\method}}(\vx,\vy^\star,\vz)\big[\underbrace{\nabla_\theta\log\pitheta(\vz|\vx)}_{\text{reasoning term}} + \underbrace{\nabla_\theta\log\pitheta(\vy^\star|\vx,\vz)}_{\text{reference answer term}}\big]\bigg].
\end{equation}
Both the objective and its gradient estimator (\Cref{eq:verifree_objective,eq:grad_ours}) are equivalent in expectation to their verifier-based counterparts (\Cref{eq:verifier_objective,eq:grad_verifier}).
Intuitively, the ``reasoning term'' in \Cref{eq:grad_ours} can be interpreted as a policy gradient where the reward for a reasoning trace $\vz$ is the probability that the policy will generate the correct answer $\vy^\star$ given $\vz$, while ``reference answer term'' can be viewed as a reward-weighted supervised learning term for $\vy^\star$ given $\vz$. We will further elaborate on this interpretation in \Cref{sec:relations}.
In addition to bypassing the need for a verifier, our \method gradient estimator also benefits from reduced variance:
\begin{theorem}(\textbf{Variance Reduction}) 
Let $\hat{G}_{\text{Verifier}}(\vx,\vy^\star,\vz)$ and $\hat{G}_{\text{\method}}(\vx,\vy^\star,\vz,\vy)$ denote the single-sample Monte Carlo estimators of $\nabla_\theta J_{\text{Verifier}}$ and $\nabla_\theta J_{\text{\method}}$ given $\vx$ and $\vy^\star$, respectively.
Then we have
\begin{equation}
\var_{\vz\sim\pitheta(\cdot|\vx)}
\big[\hat{G}_{\text{\method}}(\vx,\vy^\star,\vz)\big]
\leq
\var_{\vz\sim\pitheta(\cdot|\vx), \vy\sim\pitheta(\cdot|\vx,\vz)}
\big[\hat{G}_{\text{Verifier}}(\vx,\vy^\star,\vz,\vy)\big].
\end{equation}
\label{theorem:var_reduction}
\vspace{-4mm}
\end{theorem}
This reduction in variance arises from Rao-Blackwellization~\citep{raolackwellisation}. For intuition, the variance in the Monte Carlo estimate of $\nabla_\theta J_{\text{Verifier}}$ stems from the randomness in sampling $\vz\sim\pitheta(\cdot|\vx),\vy\sim\pitheta(\cdot|\vx,\vz)$, while for estimating $\nabla_\theta J_{\text{\method}}$ we analytically marginalizes out $\vy$, thereby removing this source of randomness. We provide a full proof in \Cref{proof:variance}.

Our gradient estimator $\nabla_\theta J_{\text{\method}}(\theta;\vx,\vy^\star)$ is fully compatible with other variance reduction techniques, including RLOO~\citep{ahmadian2024back}, GRPO~\citep{shao2024deepseekmath} reward normalizations, and the PPO~\citep{schulman2017proximal} clipping operation. As such, we sample multiple responses for each prompt and apply the RLOO baseline to the reasoning term in \Cref{eq:grad_ours}. We also adopt the corrected response-length normalization from \citet{liu2025understanding}. The final on-policy gradient estimator is as follows:
\begin{equation}
\nabla_\theta J_{\text{\method}}(\theta) =
\frac{1}{G} \sum_{i=1}^G \left[A_i\cdot\nabla_\theta\log\pitheta(\vz_i | \vx)  + R_i\cdot\nabla_\theta\log\pitheta(\vy^\star | \vx, \vz_i)\right],
\end{equation}
where $\vz_i\sim\pitheta(\cdot|\vx)$, $R_i = \pitheta(\vy^\star |\vx, \vz_i)$, and $A_i = \pitheta(\vy^\star |\vx, \vz_i) - \frac{1}{G-1}\sum_{j\neq i} \pitheta(\vy^\star |\vx, \vz_j)$.
We also provide the PPO-based off-policy variant in \Cref{appendix:ppo_grad}.

\begin{figure}[t]\small
\centering
\highlightbox{%
\begin{tabular}{@{}p{0.48\textwidth}p{0.48\textwidth}@{}}
\multicolumn{1}{c}{\textbf{Verifier-based (R1-Zero)}} & \multicolumn{1}{c}{\textbf{\method (Ours)}} \\
Model generates the \textcolor{myredd}{reasoning trace $\vz$ and answer $\vy$}. & Model generates the \textcolor{mygreenn}{reasoning trace $\vz$}. \\
\textcolor{myredd}{Extract the answer $\vy$}. & \textcolor{mygreenn}{Patch in the correct answer $\vy^\star$}. \\
\textcolor{myredd}{\textbf{Check answer using a verifier}.} & \textcolor{mygreenn}{\textbf{Evaluate probability} $\pi_\theta(\vy^\star|\vx,\vz)$.} \\
Reward \textcolor{myredd}{$R_{\text{Verifier}} = 1$ if correct, 0 otherwise.} & Reward \textcolor{mygreenn}{$R_{\text{\method}} = \pi_\theta(\vy^\star|\vx,\vz)$.} \\
Train with gradient estimator \textcolor{myredd}{$\nabla_\theta J_{\text{Verifier}}$} (Eq. \ref{eq:grad_verifier}). & Train with gradient estimator \textcolor{mygreenn}{$\nabla_\theta J_{\text{\method}}$} (Eq. \ref{eq:grad_ours}). \\
\multicolumn{2}{c}{} \\[-1em] %
\end{tabular}
}
\vspace{-3mm}
\caption{A pseudocode-like comparison of \method (ours) and the standard R1-Zero approach.}
\label{fig:pseudopseudocode}
\vspace{-.3cm}
\end{figure}

\vspace{-.1cm}
\subsection{Comparison to Existing Approaches}\label{sec:relations}
\vspace{-.1cm}

There have been two main prior works that, although derived from a different perspective, arrive at related alternative gradient estimators: \citet{tang2025learning} (JLB), and \citet{chen2024language} (LaTRO).
\begin{align*}
    \nabla_\theta J_{\text{Verifier}} &= \E_{\vz,\vy}\big[\colorbox{myred}{$\mathds{1}_{\{\vy \equiv \vy^\star\}}$}\overbrace{\nabla_\theta\log\pitheta(\vz|\vx)}^{\text{reasoning term}} + \colorbox{myblue}{$\mathds{1}_{\{\vy \equiv \vy^\star\}}$}\overbrace{\nabla_\theta\log\pitheta(\vy|\vx,\vz)}^{\text{answer term}}\big]  \tag{R1-Zero}\\
    \nabla_\theta J_{\text{\method}} &= \E_\vz\big[\colorbox{myred}{$\pitheta(\vy^\star|\vx,\vz)$}\nabla_\theta\log\pitheta(\vz|\vx) + \colorbox{myblue}{$\pitheta(\vy^\star|\vx,\vz)$}\overbrace{\nabla_\theta\log\pitheta(\vy^\star|\vx,\vz)}^{\text{reference answer term}}\big] \qquad \tag{Ours}\\
    \nabla_\theta J_{\text{JLB}} &= \E_\vz\big[\colorbox{myred}{$\log\pitheta(\vy^\star|\vx,\vz)$}\nabla_\theta\log\pitheta(\vz|\vx) + \colorbox{myblue}{$1$}\cdot\nabla_\theta\log\pitheta(\vy^\star|\vx,\vz)\big]\tag{\citet{tang2025learning}}\\
    \nabla_\theta J_{\text{LaTRO}} &= \E_\vz\big[\colorbox{myred}{$(\log\pitheta(\vy^\star|\vx,\vz)-\log\frac{\pi_\theta(\vz|\vx)}{\pi_\text{ref}(\vz|\vx)})$}\nabla_\theta\log\pitheta(\vz|\vx) + \colorbox{myblue}{$1$}\cdot\nabla_\theta\log\pitheta(\vy^\star|\vx,\vz)\big]\tag{\citet{chen2024language}}\\
\end{align*}

Both JLB and LaTRO treat the reasoning trace $\vz$ as a latent variable and extend the standard supervised learning objective (log-likelihood) to optimize lower bounds on $\log( E_{\vz\sim\pitheta(\cdot|\vx)}\left[\pitheta(\vy^\star|\vx,\vz)\right])$ and $\log( E_{\vz\sim\pi_\text{ref}(\cdot|\vx)}\left[\pitheta(\vy^\star|\vx,\vz)\right])$, respectively. The primary difference is that JLB samples $\vz$ from the learned policy $\pitheta$, while LaTRO uses a fixed reference policy $\pi_\text{ref}$. Despite originating from different perspectives, these approaches arrive at alternative gradient estimators that can be used similarly to ours, as shown in the above comparisons. However, as reported in \citet{tang2025learning}, these verifier-free, variational-inference-based methods consistently underperform the standard verifier-based R1-Zero approach. In contrast, our formulation matches or outperforms the verifier-based baseline.

One possible explanation is that our method \emph{exactly} recovers the original verifier-based objective under the single-correct-answer assumption, whereas JLB and LaTRO optimize subtly different objectives. For example, JLB effectively uses a reward of $R = \log\pitheta(\vy^\star|\vx,\vz)$, as highlighted in the gradient expressions above. Another distinction lies in the weighting of the ``reference answer term'' $\nabla_\theta\log\pitheta(\vy^\star|\vx,\vz)$. In our method, this term is weighted by the probability $\pitheta(\vy^\star|\vx,\vz)$, which is the likelihood of the reference answer given the sampled reasoning trace. In contrast, both JLB and LaTRO use a fixed weight of 1, thereby increasing the probability of $\vy^\star$ regardless of the quality of the reasoning trace $\vz$. We hypothesize that this behavior could promote poor reasoning. For instance, suppose the model generates the reasoning trace ``\texttt{...~minus 2 apples, finally resulting in a total of 7 apples}'' when the correct answer is ``\texttt{6}''. The JLB and LaTRO objectives would still push the model to output ``\texttt{6}'' from that flawed trace, reinforcing a mismatch between reasoning and answer. Our method avoids this by down-weighting contributions from low-quality traces.
\looseness=-1

\vspace{-.1cm}
\subsection{How to handle the tokenization at patching point?}
\vspace{-.1cm}
\label{subsec:split_token}
A critical consideration when extracting reasoning traces $\vz$ from model responses $(\vz, \vy)$ and subsequently replacing $\vy$ with $\vy^\star$ stems from the fact that LLMs operate on token sequences, not raw text strings. While human-readable outputs (e.g., ``\texttt{...<answer> {\textbackslash}{\textbackslash}boxed\{...\} </answer>}'' as in \Cref{template:ours}) suggest splitting reasoning traces $\vz$ at specific text patterns like ``\texttt{<answer>}'', such text-based splitting strategy may cause tokenization inconsistencies. For example, the ``\texttt{>}'' character might be tokenized differently depending on its surrounding context in $\vy$ versus $\vy^\star$.
While one potential solution is to introduce special tokens to enforce consistent tokenization boundaries, these novel tokens could harm model performance due to their absence from the base model’s pretraining vocabulary.
\looseness=-1

Instead, we resolve this by defining the end of $\vz$ at the token corresponding to \texttt{<answer}'' (i.e., without ``\texttt{>}''), leveraging the fact that the pattern ``\texttt{r>}'' does not appear in standard tokenizer vocabularies. This ensures consistent token-space alignment between sampling and optimization in RL, avoiding instability due to off-policy mismatches.
Notably, this approach is operationally equivalent to setting ``\texttt{<answer}'' (instead of ``\texttt{<answer>}'') as the stop word during sampling, a mechanism natively supported by modern inference engines like vLLM~\citep{kwon2023efficient}. 
In this case, we can sample reasoning traces $\vz$ directly, instead of first generating the full response ($\vz$, $\vy$) and then extracting $\vz$ post hoc.

%% file: sections/30-experiment.tex
\vspace{-.1cm}
\section{Experiments}
\vspace{-.1cm}
\subsection{Setup}
\vspace{-.1cm}
\textbf{Training.} Following the ``zero'' setting widely adopted in recent work~\citep{guo2025deepseek,hu2025openreasonerzeroopensourceapproach,liu2025understanding,generalreasoner}, we directly fine-tune the base LLM, skipping the intermediate stage of supervised fine-tuning (SFT). We implement our RL training pipeline using Oat~\citep{liu2024oat} by instantiating their base modules and incorporate our algorithmic changes.
Our experiments are conducted using Qwen3~\citep{qwen3} base models across multiple scales, including 1.7B, 4B, and 8B parameters. We adopt the prompt template shown in \Cref{template:ours}. 
We do not employ KL regularization losses or KL penalties in rewards, as recent studies suggest that removing KL terms does not have a significant impact~\citep{liu2025understanding,hu2025openreasonerzeroopensourceapproach}. As a result, our method does not require maintaining a reference model in memory.

\highlightbox{
\begin{template}[\textbf{for Ours}]
\label{template:ours}
\texttt{\small \\<|im\_start|>user\textbackslash n\textcolor{red}{\{question\}}\textbackslash nPlease reason step by step, and put your final answer within <answer> \textbackslash \textbackslash boxed\{\} </answer>.<|im\_end|>\textbackslash n<|im\_start|>assistant{\textbackslash}n}
\end{template}
}
For the 1.7B and 4B models, we conduct fine-tuning for approximately 4,000 policy gradient steps; for the 8B models, we fine-tune for around 3,000 policy gradient steps. During each step, the policy model (i.e., the LLM) generates 8 responses for each question (i.e., \texttt{group\_size=8}), with 16 questions processed per step. We use the sampling configurations \texttt{temperature=1.0}, \texttt{top\_p=1}, and \texttt{max\_tokens=3000} for the rollout process. The responses are then parsed into reasoning traces and model-predicted answers. We replace the model-predicted answers with the reference answers from the training dataset. Subsequently, a single forward pass is executed to compute the conditional probability of the reference answer, conditioned on all preceding tokens including the prompt and the reasoning trace. This procedure introduces only a minimal additional computational cost, as the forward pass of the LLM does not require autoregressive decoding and does not require storing intermediate states for backpropagation. All collected samples from each step are used for one optimization step. The training is conducted on a single node with 8$\times$H100 GPUs.

\textbf{Dataset.}
To support general reasoning, we begin with the dataset curated by \citet{generalreasoner}, sourced from WebInstruct~\citep{yue2024mammoth2}. To improve data quality and reliability and reduce size, we retain samples with answers that consist of fewer than seven tokens, and use Qwen2.5-72B-Instruct~\citep{qwen2.5} to filter out low-quality and noisy data. This process results in approximately 61,000 data samples spanning diverse domains, which we refer to as \textbf{WebData}. The category distribution is visualized in \Cref{fig:webdata_category_distribution}.\looseness=-1 

\textbf{Evaluation.} 
In line with prior work~\citep{generalreasoner}, we employ multiple-choice questions for evaluation to facilitate verification. To assess general reasoning abilities, we utilize the following benchmarks: \textbf{MMLU-Pro}~\citep{wang2024mmlu}, a challenging multi-task understanding benchmark designed to evaluate the capabilities of LLMs across various domains; \textbf{SuperGPQA}~\citep{du2025supergpqa}, a large-scale benchmark consisting of graduate-level reasoning questions spanning 285 diverse disciplines; and \textbf{GPQA}~\citep{rein2024gpqa}, which focuses on graduate-level question-answering and is designed to resist shallow pattern-matching and memorization.
While our primary focus is not on enhancing mathematical abilities, we also evaluate math reasoning using a suite of standard math reasoning benchmarks. This suite includes \textbf{MATH-500}~\citep{hendrycksmath2021}, \textbf{OlympiadBench}~\citep{he2024olympiadbench}, \textbf{Minerva Math}~\citep{lewkowycz2022solving}, \textbf{GSM8K}~\citep{cobbe2021training}, \textbf{AMC} and \textbf{AIME24}~\citep{li2024numinamath}. We utilize Math-Verify\footnote{\url{https://github.com/huggingface/Math-Verify}} to check for answer equivalence. Except for AIME24, where we employ a \texttt{temperature=1.0} and repeat each question 32 times,
all other benchmarks are evaluated using \texttt{temperature=0.0}.
We use \texttt{max\_tokens=8192} for all evaluations.

\begin{table}[t]
    \vspace{-.4cm}
    \fontsize{7.5}{9}\selectfont
    \tabcolsep 2.0pt
    \renewcommand{\arraystretch}{1.0}
    \caption{Accuracy comparison on the MMLU-Pro benchmark.}
    \label{tab:mmlu_pro}
    \centering
    \input{tables/mmlu_pro}
    \vspace{-.3cm}
\end{table}

\textbf{Baselines.}
Our primary baseline, denoted \textbf{Verifier}, is a verifier-based approach using the verifier from \citet{generalreasoner}.
The verifier is initialized from Qwen2.5-Math-1.5B~\citep{yang2024qwen2math} and fine-tuned on data generated by Gemini 2.0 Flash to assess equivalence between the reference and predicted answers, conditioned on the question.
We apply Dr.\,GRPO~\citep{liu2025understanding} as the optimization algorithm for the baseline, ensuring that all other settings are consistent with our approach.
Following \citet{generalreasoner}, the reward definition incorporates additional factors beyond verifier correctness, including format compliance and the length of generated answers. If the format is incorrect (e.g., missing \texttt{{\textbackslash}{\textbackslash}boxed\{\}} in the model response), a negative reward of -0.5 is applied. Moreover, a length penalty of \texttt{-0.05 $\times$ min(10, abs(length\_of\_correct\_answer - length\_of\_answer))} is added.

We also report the results for the base and the instruct models of Qwen-3-1.7B/4B/8B~\citep{qwen3} and Qwen2.5-7B~\citep{qwen2.5}, as well as the checkpoints released by Qwen2.5-7B-SimpleRL-Zoo~\citep{zeng2025simplerl}, Qwen2.5-Math-7B-Oat-Zero~\citep{liu2025understanding}, and General-Reasoner-7B~\citep{generalreasoner}.\footnote{
Qwen2.5 and Qwen3 use different naming conventions in their official release. For base models: Qwen2.5 has no suffix (e.g., Qwen2.5-7B), whereas Qwen3 adds ``-Base'' (e.g., Qwen3-8B-Base). For instruct models: Qwen2.5 uses ``-Instruct'' (e.g., Qwen2.5-7B-Instruct), while Qwen3 omits the suffix (e.g., Qwen3-8B). We follow these conventions consistently in the paper.
}
Notably, Qwen3 integrates both a \emph{thinking} mode (for complex, multi-step reasoning) and a \emph{non-thinking} mode (for rapid, context-driven responses) within a unified framework. We report results of both modes on Qwen3 instruct models.

\vspace{-.1cm}
\subsection{Main Results}
\vspace{-.1cm}

\begin{table}[t]
    \vspace{-.4cm}
    \fontsize{7.5}{9}\selectfont
    \tabcolsep 2.0pt
    \renewcommand{\arraystretch}{1.0}
    \caption{Accuracy comparison on the SuperGPQA benchmark.}
    \label{tab:superGPQA}
    \centering
    \input{tables/superGPQA}
    \vspace{-.3cm}
\end{table}

\textbf{\method{} improves general reasoning capabilities.} We begin by evaluating the effectiveness of \method{} in enhancing the general reasoning capabilities of LLMs using the MMLU-Pro and SuperGPQA benchmarks. \Cref{tab:mmlu_pro} presents a detailed comparison across model scales and domains in the MMLU-Pro benchmark. Starting from base models, applying RL with \method{} yields substantial gains in average accuracy—ranging from 12\% to 40\%—demonstrating that \method{} effectively fine-tunes LLMs to improve general reasoning performance.

Similar improvements are observed on the SuperGPQA benchmark, as shown in \Cref{tab:superGPQA}, where \method{} consistently enhances the performance of base models by a significant margin. Notably, \method{} achieves performance comparable to, or even surpassing, that of the instruct model in \textit{thinking} mode and the Verifier baseline (i.e., the RL-tuned model learned with an additional model-based verifier)—without relying on any explicit verification signals.

In addition to accuracy gains, we also observe an increase in response length after tuned by \method{}, suggesting that the model explores longer reasoning traces to arrive at more accurate answers—a behavior reminiscent of DeepSeek-R1-Zero~\citep{guo2025deepseek}. Results on the GPQA benchmark are provided in \Cref{app:more_results} due to space constraints.

\begin{figure*}[t]
\vspace{-.1cm}
\centering
\includegraphics[width=.9\textwidth]{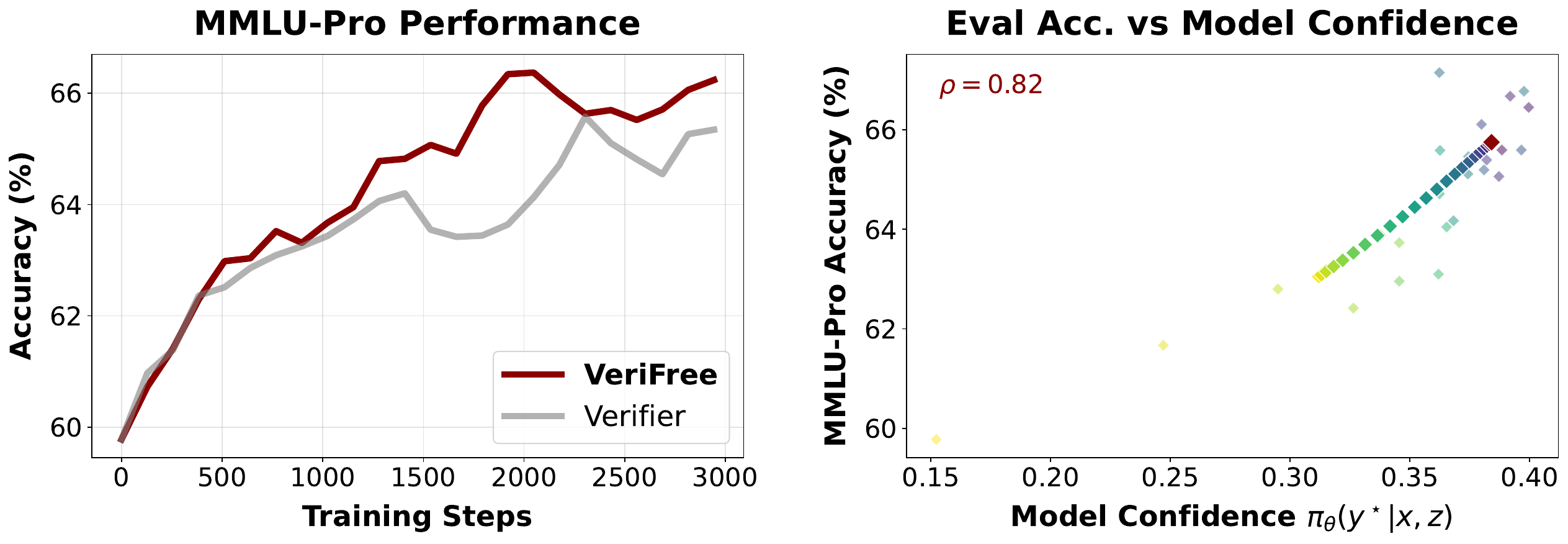}
  \caption{\textbf{Left}: MMLU-Pro accuracy of \method and the baseline fine-tuned from Qwen3-8B base model along training steps. The curve is smoothed by a moving average with an interval of 384. \textbf{Right}: The dynamics of MMLU-Pro evaluation accuracy and average model confidence $\pitheta(\vy^\star |\vx, \vz)$ along training based on Qwen3-8B base model. Raw data points are depicted with more transparency, while smoothed data using a Gaussian filter is shown with less transparency for emphasis. Darker colors represent larger training steps.}{\label{fig:mmlu_pro_accuracy_curve}}
  \vspace{-.3cm}
\end{figure*}

\textbf{\method{} leads to better learning efficiency.} We compare \method{} with the baseline that learns from a model-based verifier reward (i.e., Verifier). As shown in \Cref{fig:mmlu_pro_accuracy_curve}\,(Left), \method{} consistently outperforms the baseline, achieving higher accuracy with fewer training steps. We attribute this improved learning efficiency to reduced gradient variance, enabled by \method{}’s 
continuous reward signals
and the RLOO objective. While both approaches optimize the same reward signal in expectation, \method{} provides more stable and informative policy gradients, which accelerate convergence and leads to better final performance.

\textbf{Model confidence is a good reasoning capability proxy.} 
Our analysis based on Qwen3-8B base model reveals a strong positive correlation ($\rho = 0.82$) between MMLU-Pro accuracy and the average model confidence $\pitheta(\vy^\star|\vx, \vz)$ during training (\Cref{fig:mmlu_pro_accuracy_curve}, Right). This empirically demonstrates that the model's self-estimated confidence in the correct answer, i.e., $\pitheta(\vy^\star|\vx, \vz)$, serves as an effective metric for quantifying emergent reasoning capabilities in LLMs.

\begin{wrapfigure}{r}{0.55\textwidth}
    \vspace*{-0.5cm}
    \centering
    \includegraphics[width=0.47\textwidth]{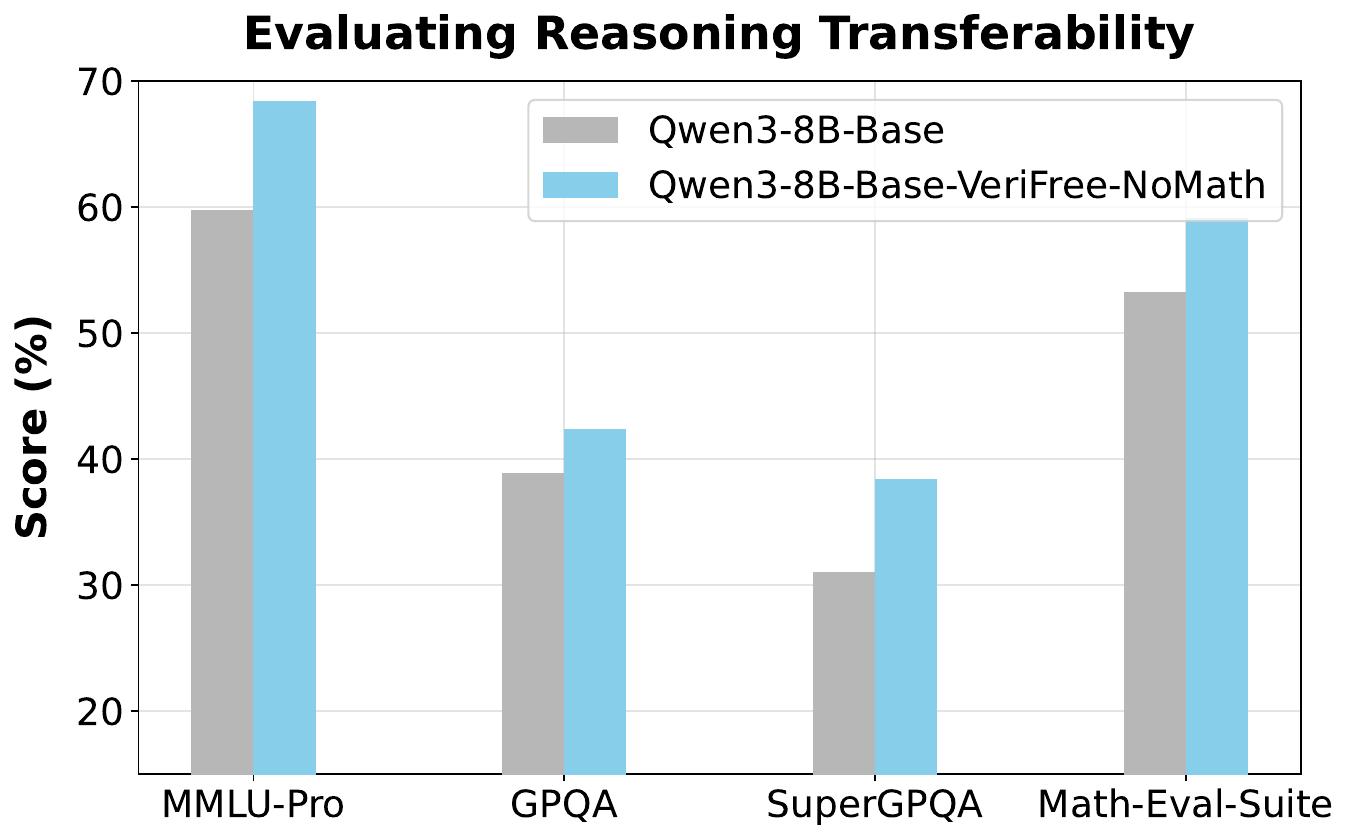}
    \vspace{-0.2cm}
    \caption{Reasoning transfer to math without math training. \method{} enhances reasoning transfer: when trained only on non-math data, the model improves on general benchmarks and effectively transfers to math-specific tasks.}
    \label{fig:no_math}
    \vspace*{-.2cm}
\end{wrapfigure}

\textbf{\method{} learns transferable reasoning skills.} To evaluate the transferability of reasoning acquired through \method{}, we train a model on a dataset with all math-related examples removed, and assess its performance on both general and math-specific benchmarks. As shown in \Cref{fig:no_math}, \method{} not only improves reasoning performance on general tasks, as expected, but also demonstrates strong transfer to math benchmarks—despite the absence of math supervision during training. This highlights \method{}'s ability to induce general reasoning capabilities that extend across domains.

\vspace{-.1cm}
\subsection{Ablation Study}
\vspace{-.1cm}
To systematically evaluate method components and offer a comprehensive understanding of \method, we conduct ablation studies based on Qwen3-1.7B base models as follows. 

\begin{figure*}[t]
\centering
\includegraphics[width=.9\textwidth]{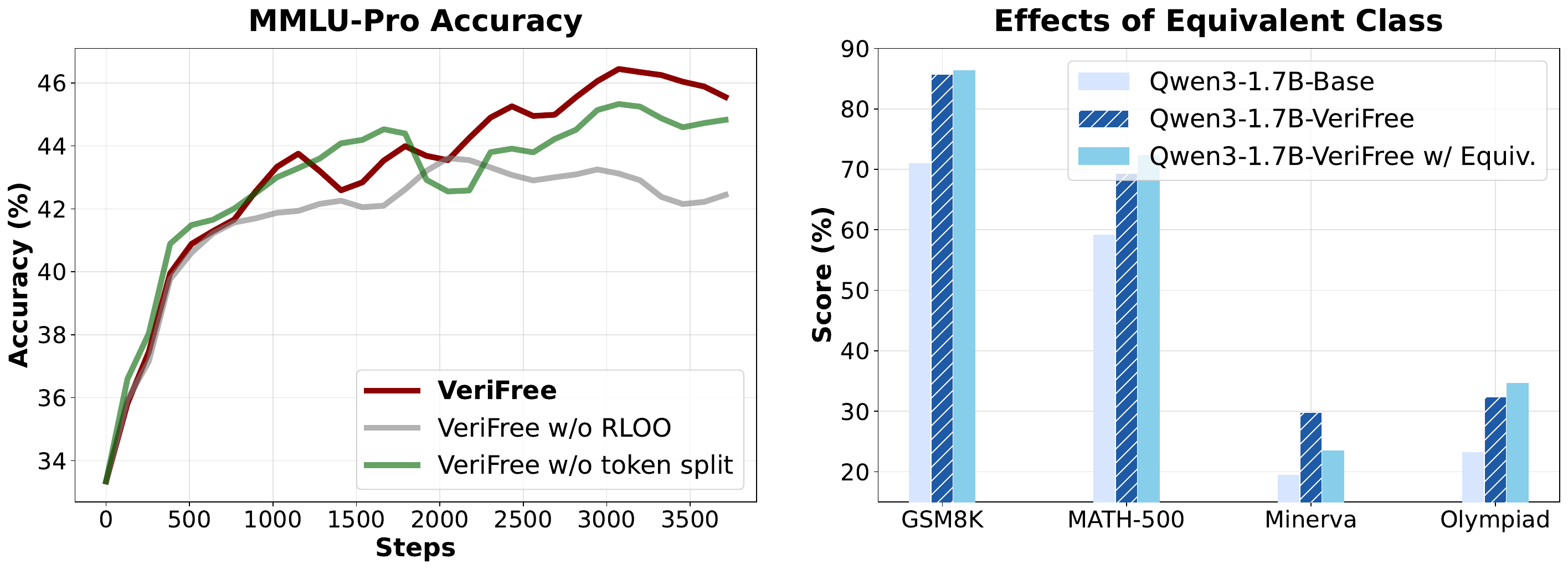}
  \caption{\textbf{Left}: MMLU-Pro Evaluation Accuracy of \method against ablation variants (w/o RLOO, w/o tokenization-aware split strategy) along training steps. Note that all these models are based on Qwen3-1.7B base models. \textbf{Right}: Effects of introducing the equivalent class to \method on model performance.
}{\label{fig:ablation}}
\vspace{-.3cm}
\end{figure*}

\textbf{Effects of extraction strategy for reasoning trace $\vz$.}  
Our method requires precise separation between reasoning path $\vz$ and answer $\vy$ to enable answer replacement. While human-readable splits using \texttt{<answer>} seem intuitive, we instead define $\vz$ to end at ``\texttt{<answer}'' (omitting ``\texttt{>}''), ensuring consistent tokenization boundaries (see \Cref{subsec:split_token}). 
We compare with a variant using text-based splitting (denoted as ``\method\ w/o token split'') on Qwen3-1.7B via MMLU-Pro (\Cref{fig:ablation}). Our tokenization-aware approach achieves superior convergence, while the variant suffers optimization instability due to effectively introducing off-policy data.

\textbf{Effects of RLOO.} As observed in \Cref{fig:ablation} (Left), removing RLOO leads to a consistent drop in performance throughout training, with final accuracy more than $3\%$ lower than that of the full method. This highlights the importance of RLOO in stabilizing learning and guiding the model toward better generalization. Without RLOO, the model converges prematurely and fails to reach the same peak accuracy.
\looseness=-1

\textbf{Effects of equivalence class.}
As mentioned in \Cref{subsec:method}, verifier-based RL typically assesses answer correctness as rewards. Correct answers within a specific class often form an equivalence class. Our method, however, utilizes model confidence by focusing on a single reference answer for a given question and the model's reasoning trace. To explore the potential advantages of integrating an ``equivalence class'' into our approach, we conducted ablation studies as follows. We employed a model fine-tuned on the MATH-12k dataset~\citep{hendrycks2021measuring,lightman2024lets} from Qwen3-8B base model through Dr.\,GRPO~\citep{liu2025understanding} with rule-based verification to sample answers on MATH-12k, subsequently verifying answer correctness using Math-Verify. This approach enabled us to create an extended dataset with a set of equivalent correct answers for each question. We then fine-tuned Qwen3-1.7B base models using our method on both the original and the extended MATH-12k datasets incorporating equivalence classes. These models are evaluated on GSM8K, MATH-500, Minerva Math, and OlympiadBench to assess the impact of including equivalence classes. The results, shown in \Cref{fig:ablation} (Right), indicate that considering equivalence classes in our method offers slight performance improvements, aligning with our expectations. This highlights a minor limitation of our current formulation and motivates future work on algorithms that can better leverage answer equivalence.

%% file: tables/mmlu_pro.tex
\begin{tabular}{l|c|c|cccccccccccccc}
\toprule
Method & Len. & Avg. & CS & Math & Chem & Eng & Law & Bio & Health & Phys & Bus & Phil & Econ & Other & Psy & Hist
\\

\midrule

Qwen3-1.7B-Base	&618&	33.3	&	34.6	&38.9	&32.2	&21.0	&17.3	&56.1	&33.5	&32.0	&38.5	&21.8	&45.7	&28.4	&44.4	&21.0 \\
Qwen3-1.7B (w/o thinking)	&893		&46.1		&49.5	&64.4	&48.0	&35.9	&22.9	&64.9	&38.0	&49.7	&53.5	&33.7	&53.9	&36.4	&51.6	&31.2 \\
Qwen3-1.7B (w/ thinking)	&3904		&52.0		&56.1	&76.4	&57.6	&27.0	&21.9	&67.9	&47.7	&57.5	&61.3	&38.9	&64.5	&42.5	&59.2	&32.3 \\
Qwen3-1.7B-Base-Verifier	&875		&\textbf{47.0}		&48.8	&64.4	&52.7	&38.1	&18.7	&62.9	&41.2	&51.9	&54.9	&31.9	&55.2	&38.6	&53.3	&30.2 \\
\rowcolor{table-blue!66} Qwen3-1.7B-Base-\method	&856		&\underline{46.9}		&46.8	&64.1	&51.7	&41.8	&20.0	&64.0	&39.7	&52.1	&55.6	&29.5	&53.1	&37.5	&53.0	&29.9		\\										
\midrule
Qwen3-4B-Base	&825		&47.2		&42.9	&67.1	&55.5	&40.0	&22.5	&56.9	&43.6	&55.4	&54.9	&27.5	&52.7	&34.3	&48.6	&34.7	\\											
Qwen3-4B (w/o thinking)	&838		&60.0		&65.9	&79.1	&65.8	&45.7	&29.0	&76.6	&57.0	&65.1	&66.7	&48.9	&69.2	&52.1	&64.3	&44.6	\\											
Qwen3-4B (w/ thinking)	&3456		&62.7		&70.0	&84.8	&66.6	&38.6	&28.7	&81.3	&60.4	&67.4	&69.2	&53.7	&75.1	&57.8	&67.9	&49.6 \\

Qwen3-4B-Base-Verifier	&921		&\underline{63.0}		&66.1	&81.3	&69.7	&52.8 &29.1	&79.8	&62.8	&67.6	&71.2	&48.5	&73.1	&52.8	&68.5	&45.4		\\										

\rowcolor{table-blue!66}Qwen3-4B-Base-\method	&1241		&\textbf{63.5}		&64.4	&82.2	&70.1	&55.6	&30.7	&81.7	&59.2	&71.0	&71.0	&47.1	&71.7	&53.4	&66.8	&47.5	\\

\midrule
Qwen2.5-7B  & 519	&	47.8	&	48.3	&59.5	&44.4	&33.4	&25.1	&63.6	&50.4	&48.0	&55.9	&34.7	&60.6	&46.0	&58.2	&38.3 \\
Qwen2.5-7B-SimpleRL-Zoo  &705	&51.2	&51.2	&52.0	&50.2	&40.8	&30.5	&69.5	&54.3	&52.5	&57.3	&41.9	&62.8	&52.6	&60.8	&42.3 \\
Qwen2.5-Math-7B-Oat-Zero &556 &40.5	&47.6	&47.7	&46.9	&32.1	&18.1	&53.6	&25.7	&49.4	&52.9	&29.5	&54.7	&32.8	&43.0	&22.8 \\
Qwen2.5-7B-Instruct  &481		&55.3		&56.6	&70.4	&55.6	&42.7	&29.8	&69.3	&55.1	&57.9	&63.5	&41.5	&63.4	&53.6	&62.4	&43.6 \\

General-Reasoner-7B &867		&58.7		&63.4	&73.7 &63.3	&44.9	&35.2	&72.0	&56.6	&61.5	&66.7	&43.1	&68.1	&52.8	&62.8	&47.8 \\
Qwen3-8B-Base &613		&59.8		&61.2	&75.0	&66.2	&46.7	&31.4	&75.9	&60.4	&62.1	&65.9	&48.7	&69.0	&54.3	&63.9	&47.2 \\
Qwen3-8B (w/o thinking) &1032		&61.9		&65.6	&71.9	&62.8	&46.2	&34.7	&79.9	&66.1	&63.7	&69.3	&55.9	&72.9	&58.9	&67.9	&52.5 \\
Qwen3-8B (w/ thinking) &3952		&\underline{66.9}		&71.5	&83.8	&68.0	&38.7	&39.2	&85.2	&72.1	&69.8	&73.3	&57.5	&79.2	&66.3	&71.8	&57.7 \\
Qwen3-8B-Base-Verifier	&594		&65.9		&63.9	&81.8	&71.1	&56.9	&35.4	&81.9	&64.9	&71.6	&74.1	&53.9	&74.2	&58.4	&68.4	&54.3	\\											

\rowcolor{table-blue!66}Qwen3-8B-Base-\method  &  776 & \textbf{67.2} & 71.5 & 85.3 & 73.5 & 55.7 & 37.3	&81.9	&64.3	&73.1	&74.1	&54.9	&74.8	&59.6	&67.7	& 54.6 \\
\bottomrule
\end{tabular}

%% file: tables/superGPQA.tex
\begin{tabular}{l|c|c|ccccccccccccc}
\toprule
Method & Len. & Avg. & Eng. & Med. & Sci. & Phil. & M.S. & Econ. & Mgmt. & Socio. & L/A & Hist. & Agron. & Law & Edu.
\\
\midrule
Qwen3-1.7B-Base	&997		&17.4	&17.7	&18.6	&16.0	&27.4	&27.3	&20.5	&22.4	&23.1	&15.5	&11.1	&18.1	&20.6	&21.3 \\
Qwen3-1.7B (w/o thinking)	&1152		&23.3	&22.6	&22.8	&24.3	&30.3	&31.2	&24.3	&27.5	&23.8	&19.0	&18.1	&20.8	&24.5	&28.1 \\
Qwen3-1.7B (w/ thinking)	&4799		&23.6	&21.8	&25.3	&23.6	&33.1	&33.7	&29.6	&27.5	&32.2	&19.0	&18.0	&25.0	&26.2	&31.6 \\
Qwen3-1.7B-Base-Verifier 	&1049		&\underline{24.5}	&26.0	&23.9	&24.4	&30.8	&26.8	&26.9	&27.0	&26.6	&18.9	&16.6	&22.3	&22.6	&27.3	\\	
\rowcolor{table-blue!66}Qwen3-1.7B-Base-VeriFree	&964		&\textbf{24.8}	&25.7	&24.7	&24.9	&26.5	&30.2	&25.9	&27.9	&28.0	&20.4	&15.9	&22.9	&25.0	&28.9	\\								
\midrule
Qwen3-4B-Base	&902	&24.7	&25.7	&23.6	&26.0	&23.6	&25.4	&28.8	&28.4	&19.6	&16.4	&16.8	&20.6	&25.6	&24.0	\\		
Qwen3-4B (w/o thinking)	&1397		&31.6	&32.0	&31.5	&32.3	&37.5	&36.1	&37.8	&33.7	&33.6	&24.3	&20.6	&28.3	&31.4	&33.5 \\
Qwen3-4B (w/ thinking)	&4568		&31.7	&30.7	&33.2	&32.1	&41.2	&31.7	&41.7	&35.9	&32.9	&24.5  &22.4	&30.9	&35.7	&35.3 \\
Qwen3-4B-Base-Verifier	&1045	&\underline{34.3}	&35.4 &35.5	&34.5	&39.2	&41.0	&39.1	&36.7	&37.1	&26.6	&22.3	&33.8	&33.1	&35.3				\\					
\rowcolor{table-blue!66}Qwen3-4B-Base-VeriFree	&1451		&\textbf{35.1}	&36.3	&34.5	&36.9	&35.7	&37.1	&39.1	&38.3	&31.5	&24.7	&22.0	&33.0	&33.2	&34.1 \\	

\midrule
Qwen2.5-7B	&716		&23.8	&24.2	&27.0	&21.8	&28.8	&31.2	&27.6	&29.1	&22.4	&20.8	&20.2	&24.5	&27.4	&30.2 \\
Qwen-2.5-7B-SimpleRL-Zoo	&850		&26.3	&26.4	&30.5	&23.8	&32.6	&32.2	&33.0	&31.9	&28.7	&24.1	&21.4	&27.2	&29.6	&32.9\\
Qwen2.5-Math-7B-Oat-Zero	&638		&21.3	&23.1	&16.4	&21.5	&23.1	&21.5	&25.9	&27.2	&25.2	&17.7	&15.9	&21.4	&18.8	&24.8\\
Qwen2.5-7B-Instruct	&604		&28.4	&27.7	&32.2	&27.6	&33.7	&32.2	&32.4	&32.9	&32.9	&24.5	&22.1	&29.7	&30.6	&32.4\\
General-Reasoner-7B &1047		&30.8	&31.5	&32.2	&29.9	&35.2	&41.5	&38.4	&33.1	&35.0	&25.5	&22.7	&28.9	&32.5	&35.5		\\	
Qwen3-8B-Base & 825		&31.0	&31.3	&34.0	&30.6	&36.0	&37.1	&34.7	&37.5	&35.0	&24.2	&20.0	&28.5	&31.4	&36.4\\
Qwen3-8B (w/o thinking)	&1638		&32.4	&32.6	&36.5	&31.2	&39.5	&42.0	&37.7	&37.3	&38.5	&25.0	&22.6	&33.2	&34.3	&38.4\\
Qwen3-8B (w/ thinking)	&4995		&35.0	&33.6	&42.1	&33.5	&44.4 &37.6	&44.2	&42.7	&42.7	&27.9	&24.9	&37.9	&38.7	&40.7\\
Qwen3-8B-Base-Verifier	&713		&\underline{37.1}	&38.2	&39.5	&37.2	&39.5	&39.5	&43.0	&40.1	&38.5	&28.9	&24.8	&34.2	&34.8	&38.2		\\							
\rowcolor{table-blue!66}Qwen3-8B-Base-VeriFree    & 951 & \textbf{38.0} & 38.3&	39.1	&39.6&	37.5	&42.9	&41.8	&41.7	&44.8	&28.6	&23.3	&33.6	&36.3	&38.6\\
\bottomrule
\end{tabular}

%% file: sections/40-related.tex
\vspace{-.1cm}
\section{Related Works}
\vspace{-.1cm}
\textbf{DeepSeek-R1-Zero-like reinforcement learning.} DeepSeek-R1-Zero~\citep{guo2025deepseek}, T\"ulu \citep{lambert2024tulu3} and OpenAI's o1 \citep{o1} recently demonstrated that applying RL to learn directly from binary verification-based rewards can be extremely powerful in enhancing the reasoning capabilities of base LLMs. Since then several works have reproduced R1-Zero-like training on smaller scales \citep{zeng2025simplerl,tinyzero,liu2025there,hu2025openreasonerzeroopensourceapproach,deepscaler2025,deepcoder2025}. The aforementioned works all focus on math and coding, domains where verifiable rewards are readily available. By contrast, our work aims to extend the R1-Zero-like training paradigm to general domains where verifiable rewards are not available.

\textbf{Reasoning beyond verifiable domains.}
Previous work on reasoning without access to verifiable rewards has been based around employing an additional LLM to act as a proxy verifier or reward model. NaturalReasoning~\citep{yuan2025naturalreasoning} introduces a large, multi-domain dataset and presents baselines trained using RFT \citep{yuan2023scalingrft} and DPO \citep{rafailov2023direct}, leveraging a second LLM as a reward model, while \citet{su2025expanding} and General-Reasoner~\citep{generalreasoner} similarly incorporate a separate LLM to serve as a verifier. X-Reasoner~\citep{liu2025xreasonergeneralizablereasoningmodalities} also investigates general reasoning but circumvents the lack of rule-based verification by dropping the R1-Zero-style training paradigm, instead training via SFT~\citep{grattafiori2024llama} on responses sampled from more capable models.
\looseness=-1

\textbf{Self-improving language models.} Several works have explored training LLMs using signals based on the model's own outputs. \citet{yuan2024selfrewarding} propose to prompt the model to judge and rank different responses and select the best and worst for preference learning. \citet{chen2025bootstrapping} leverage the DPO implicit rewards for more efficient and robust self-alignment via iterative DPO~\citep{rafailov2023direct}. \citet{zuo2025ttrl} use majority voting to construct self-labeled rewards for RL to further improve well-trained models during test time, which can be understood as a form of sharpening~\citep{huang2024self}. Another line of research~\citep{phan2023training,chen2024language,tang2025learning} approaches LLM reasoning from the direction of variational optimization, treating the reasoning trace as a latent variable. Despite starting from a different viewpoint our method has interesting and close connections to this perspective which we discuss in detail in \Cref{sec:relations}.

%% file: sections/50-conclusion.tex
\vspace{-.1cm}
\section{Conclusions}
\vspace{-.1cm}
In this paper, we rethink reinforcement learning with verifiable rewards (RLVR) for LLMs from a novel perspective.
By leveraging the gradient equivalence under the unique answer assumption, we derive a new optimization objective that eliminates the need for explicit verification.
Our proposed method, \method{}, is particularly well-suited for general reasoning tasks, where rule-based verifiers are infeasible and model-based verifiers are both expensive and vulnerable to reward hacking.
Through extensive experiments and ablations, we demonstrate the effectiveness and robustness of \method{} on a wide range of general reasoning benchmarks. We hope our work offers a fresh viewpoint for the LLM RL community and provides a practical approach for building future general-purpose reasoners.

%% file: sections/appendix/A00-related.tex
\clearpage
\appendix

\section*{\LARGE Appendix}

\vspace*{20pt}
\section*{Table of Contents}
\vspace*{-5pt}
\startcontents[sections]
\printcontents[sections]{l}{1}{\setcounter{tocdepth}{2}}

%% file: sections/appendix/A13-proofs.tex
\section{Theoretical Analysis}

\subsection{Derivation of Gradient Estimators}
Here we provide derivations for \Cref{eq:grad_ours}---the gradient estimator of \method. We also include the corresponding derivation for the standard verifier-based gradient estimator for completeness. 
\begin{proof}\label{proof:gradients}
The gradient estimator for $J_{\text{Verifier}}$ is derived as follows:
\begin{align*}
    & \nabla_\theta J_{\text{Verifier}}(\theta;\vx,\vy^\star) 
    \\
    =\, &\nabla_\theta\E_{\vz\sim\pi_\theta(\cdot|\vx)}
    \Big[
        \E_{\vy\sim\pi_\theta(\cdot|\vx,\vz)}
        \big[ 
            R_{\text{Verifier}}(\vy;\vy^\star)
        \big]
    \Big]
    \\
    =\, &\nabla_\theta\sum_{\vz,\vy}R_{\text{Verifier}}(\vy;\vy^\star)\pi_\theta(\vy|\vx,\vz)\pi_\theta(\vz|\vx)
    \\
    =\, &\sum_{\vz,\vy}R_{\text{Verifier}}(\vy;\vy^\star)
    \big[\pi_\theta(\vy|\vx,\vz)\nabla_\theta\pi_\theta(\vz|\vx) + \pi_\theta(\vz|\vx)\nabla_\theta\pi_\theta(\vy|\vx,\vz)\big]
    \\
    =\, &\sum_{\vz,\vy}R_{\text{Verifier}}(\vy;\vy^\star)\big[\pi_\theta(\vy|\vx,\vz)\pi_\theta(\vz|\vx)\nabla_\theta\log\pi_\theta(\vz|\vx) + \pi_\theta(\vz|\vx)\pi_\theta(\vy|\vx,\vz)\nabla_\theta\log\pi_\theta(\vy|\vx,\vz)\big]
    \\
    =\, &\E_{\vz\sim\pi_\theta(\cdot|\vx)}
    \bigg[\E_{\vy\sim\pi_\theta(\cdot|\vx,\vz)}\Big[R_{\text{Verifier}}(\vy;\vy^\star)
    \big[ 
        \nabla_\theta\log\pi_\theta(\vz|\vx) + \nabla_\theta\log\pi_\theta(\vy|\vx,\vz) 
    \big] \Big]\bigg]
\end{align*}
The gradient estimator for $J_{\text{\method}}$  is derived as follows:
\begin{align*}
    & \nabla_\theta J_{\text{\method{}}}(\theta;\vx,\vy^\star) 
    \\
    =\, &\nabla_\theta\E_{\vz\sim\pi_\theta(\cdot|\vx)}
    [R_{\text{\method{}}}(\vz;\vx,\vy^\star)]
    \\
    =\, &\nabla_\theta\E_{\vz\sim\pi_\theta(\cdot|\vx)}
    [\pi_\theta(\vy^\star|\vx,\vz)]
    \\
    =\, &\nabla_\theta\sum_{\vz}\pi_\theta(\vy^\star|\vx,\vz)\pi_\theta(\vz|\vx)
    \\
    =\, &\sum_{\vz}
    \big[
        \pi_\theta(\vy^\star|\vx,\vz)\nabla_\theta\pi_\theta(\vz|\vx) + \pi_\theta(\vz|\vx)\nabla_\theta\pi_\theta(\vy^\star|\vx,\vz)
    \big]
    \\
    =\, &\sum_{\vz} \big[ \pi_\theta(\vy^\star|\vx,\vz)\pi_\theta(\vz|\vx)\nabla_\theta\log\pi_\theta(\vz|\vx) + \pi_\theta(\vz|\vx)\pi_\theta(\vy^\star|\vx,\vz)\nabla_\theta\log\pi_\theta(\vy^\star|\vx,\vz)\big]\\
    =\, &\E_{\vz\sim\pi_\theta(\cdot|\vx)}
    \Big[\pi_\theta(\vy^\star|\vx,\vz) \big[\nabla_\theta\log\pi_\theta(\vz|\vx) + \nabla_\theta\log\pi_\theta(\vy^\star|\vx,\vz) \big]\Big]
    \\
    =\, &\E_{\vz\sim\pi_\theta(\cdot|\vx)}
    \Big[ 
    R_{\text{\method{}}}(\vz;\vx,\vy^\star)
        \big[ \nabla_\theta\log\pi_\theta(\vz|\vx) + \nabla_\theta\log\pi_\theta(\vy^\star|\vx,\vz) \big]
    \Big]
\end{align*}
\end{proof}

\subsection{Proof of Lower Variance}
\label{proof:variance}
Here we provide a full proof of \Cref{theorem:var_reduction}, the reduced variance property of \method.
We show that the policy gradient estimator derived from $J_{\text{\method{}}}(\theta;\vx,\vy^\star)$ has variance less than or equal to that of the estimator derived from $J_{\text{Verifier}}(\theta;\vx,\vy^\star)$ for any given $\vx,\vy^\star$. The same relationship will hold for the global objectives averaged over $(\vx,\vy^\star) \sim \mathcal{D}$.

\begin{proof}
The global objective functions are:
\begin{align*}
    J_{\text{Verifier}}(\theta) &= \E_{(\vx,\vy^\star)\sim\mathcal{D}} \left[ \E_{\vz\sim\pi_\theta(\cdot|\vx)} \Big[ \E_{\vy\sim\pi_\theta(\cdot|\vx,\vz)}\big[\mathbbm{1}\{\vy=\vy^\star\}\big] \Big] \right] 
    \\
    J_{\text{\method{}}}(\theta) &= \E_{(\vx,\vy^\star)\sim\mathcal{D}} \Big[ \E_{\vz\sim\pi_\theta(\cdot|\vx)}\big[\pi_\theta(\vy^\star|\vx,\vz) \big] \Big]
\end{align*}
For a given $(\vx,\vy^\star) \sim \mathcal{D}$, the single-sample Monte Carlo gradient estimators are:
\begin{align*}
\hat{G}_{\text{Verifier}}(\vx,\vy^\star,\vz,\vy) &= \mathbbm{1}\{\vy=\vy^\star\} \big[ \nabla_\theta \log \pi_\theta(\vz|\vx) + \nabla_\theta \log \pi_\theta(\vy|\vx,\vz) \big]  
\\
\intertext{where $\vz \sim \pi_\theta(\cdot|\vx)$, $\vy \sim \pi_\theta(\cdot|\vx,\vz)$, and}
\hat{G}_{\text{\method{}}}(\vx,\vy^\star,\vz) &= \pi_\theta(\vy^\star|\vx,\vz) \big[ \nabla_\theta \log \pi_\theta(\vz|\vx) + \nabla_\theta \log \pi_\theta(\vy^\star|\vx,\vz) \big]  
\\
\intertext{where $\vz \sim \pi_\theta(\cdot|\vx)$.}
\end{align*}
The proof relies on the law of total variance and the relationship between $\hat{G}_{\text{Verifier}}$ and $\hat{G}_{\text{\method{}}}$.

First, we show that $\hat{G}_{\text{\method{}}}(\vx,\vy^\star,\vz)$ is the conditional expectation of $\hat{G}_{\text{Verifier}}(\vx,\vy^\star,\vz,\vy)$ given $\vx, \vy^\star, \vz$. The expectation is taken over $\vy \sim \pi_\theta(\cdot|\vx,\vz)$:
\begin{align*}
    &\E_{\vy \sim \pi_\theta(\cdot|\vx,\vz)} 
    \big[ 
        \hat{G}_{\text{Verifier}}(\vx,\vy^\star,\vz,\vy) | \vx,\vy^\star,\vz 
    \big] 
    \\
    =\, &\E_{\vy \sim \pi_\theta(\cdot|\vx,\vz)} 
    \Big[ 
        \mathbbm{1}\{\vy=\vy^\star\} 
            \big[ \nabla_\theta \log \pi_\theta(\vz|\vx) + \nabla_\theta \log \pi_\theta(\vy|\vx,\vz) \big]
        \Big| \vx,\vy^\star,\vz 
    \Big] 
    \\
    =\, &\sum_{\vy'} \pi_\theta(\vy'|\vx,\vz) 
    \Big[ 
        \mathbbm{1}\{\vy'=\vy^\star\} \big[ \nabla_\theta \log \pi_\theta(\vz|\vx) + \nabla_\theta \log \pi_\theta(\vy'|\vx,\vz) \big] 
    \Big] 
    \\
    =\, &\pi_\theta(\vy^\star|\vx,\vz) 
        \big[ 
            \nabla_\theta \log \pi_\theta(\vz|\vx) + \nabla_\theta \log \pi_\theta(\vy^\star|\vx,\vz) 
        \big] 
    \\
    =\, &\hat{G}_{\text{\method{}}}(\vx,\vy^\star,\vz)
\end{align*}
We denote $\E_{\vy \sim \pi_\theta(\cdot|\vx,\vz)} 
    \big[ 
        \hat{G}_{\text{Verifier}}(\vx,\vy^\star,\vz,\vy) \big| \vx,\vy^\star,\vz 
    \big]$ 
as $\E_{\vy|\vz} \big[ \hat{G}_{\text{Verifier}}(\vx,\vy^\star,\vz,\vy) \big]$ for brevity in the following, since $\vx$ and $\vy^\star$ are already given and fixed. Thus, we have 
\begin{align}
    \E_{\vy|\vz} \big[ \hat{G}_{\text{Verifier}}(\vx,\vy^\star,\vz,\vy) \big]
    = \hat{G}_{\text{\method{}}}(\vx,\vy^\star,\vz)
    \label{eq:proof_step1}
\end{align}

The law of total variance states that for a random variable $W$ and conditioning variables $S$, $\var(W) = \E_S[\var[W|S]] + \var_S[\E[W|S]]$.
Let $W = \hat{G}_{\text{Verifier}}(\vx,\vy^\star,\vz,\vy)$. 
Given $\vx$ and $\vy^\star$, the sources of randomness for $\hat{G}_{\text{Verifier}}$ are $\vz$ and $\vy$.
Let $S = \vz$ be the conditioning variables. The randomness in $\hat{G}_{\text{Verifier}}$ given $S$ comes from $\vy \sim \pi_\theta(\cdot|\vx,\vz)$.
Applying the law:
\begin{align*}
    \var_{\vz,\vy} \big[ \hat{G}_{\text{Verifier}}(\vx,\vy^\star\!\!,\vz,\vy) \big]
    = 
    \E_{\vz} \! \Big[ \var_{\vy|\vz} \big[ \hat{G}_{\text{Verifier}}(\vx,\vy^\star\!\!,\vz,\vy) \big] \Big] 
    \! + \! 
    \var_{\vz} \! \Big[ \E_{\vy|\vz} \big[ \hat{G}_{\text{Verifier}}(\vx,\vy^\star\!\!,\vz,\vy) \big] \Big]
\end{align*}
The expectation $\E_{\vz}$ is over $\vz \sim \pi_\theta(\cdot|\vx)$. The conditional variance $\var_{\vy|\vz}$ and expectation $\E_{\vy|\vz}$ are over $\vy \sim \pi_\theta(\cdot|\vx,\vz)$ for fixed $\vx,\vy^\star,\vz$.

Substituting the result from \cref{eq:proof_step1}
into the law of total variance:
\begin{align*}
    \var_{\vz,\vy} \big[ \hat{G}_{\text{Verifier}}(\vx,\vy^\star\!\!,\vz,\vy) \big] 
    = 
    \E_{\vz} \! \Big[ \var_{\vy|\vz} \big[ \hat{G}_{\text{Verifier}}(\vx,\vy^\star\!\!,\vz,\vy) \big] \Big] 
    + 
    \var_{\vz} 
    \big[ 
        \hat{G}_{\text{\method{}}}(\vx,\vy^\star,\vz) 
    \big]
\end{align*}
The second term, $\var_{\vz} \big[ \hat{G}_{\text{\method{}}}(\vx,\vy^\star,\vz) \big]$, is the definition of the variance of the estimator $\hat{G}_{\text{\method{}}}$.
The first term, $\E_{\vz} \Big[ \var_{\vy|\vz} \big[ \hat{G}_{\text{Verifier}}(\vx,\vy^\star,\vz) \big] \Big]$, is an expectation of a variance. Since variance is always non-negative, $\var_{\vy|\vz} \big[ \hat{G}_{\text{Verifier}}(\vx,\vy^\star,\vz) \big] \ge 0$. Therefore, its expectation is also non-negative:
\begin{align*}
\E_{\vz} \Big[ \var_{\vy|\vz} \big[ \hat{G}_{\text{Verifier}}(\vx,\vy^\star,\vz) \big] \Big]
\ge 0
\end{align*}
Thus, we have:
\begin{align*}
\var_{\vz,\vy} \big[ \hat{G}_{\text{Verifier}}(\vx,\vy^\star\!\!,\vz,\vy) \big]
= 
(\text{a non-negative term}) 
+
\var_{\vz} 
    \big[ 
        \hat{G}_{\text{\method{}}}(\vx,\vy^\star,\vz) 
    \big]
\end{align*}
This implies:
\begin{align*}
\var_{\vz,\vy} \big[ \hat{G}_{\text{Verifier}}(\vx,\vy^\star\!\!,\vz,\vy) \big]
\ge 
\var_{\vz} 
    \big[ 
        \hat{G}_{\text{\method{}}}(\vx,\vy^\star,\vz) 
    \big]
\end{align*}
The variance of the policy gradient estimator $\hat{G}_{\text{\method{}}}$ is less than or equal to that of $\hat{G}_{\text{Verifier}}$. This is an instance of Rao-Blackwellization, where analytically integrating out a source of randomness (the sampling of $\vy$) by using its conditional expectation reduces variance.
\end{proof}

\section{Off-policy Gradient Estimators}
\label{appendix:ppo_grad}

In the main paper we provide an expression for the gradient estimator when the data is fully on-policy. \method is also fully compatible with PPO-style gradient clipping for the case when data is reused to improve sample efficiency. In this case the gradient estimator is:

\begin{align*}
\nabla_\theta J_{\method{}}(\theta) & =
\frac{1}{G} \sum_{i=1}^G \Bigg[\sum_{t=1}^{\left|\vz_i\right|}\text{Clip}\bigg\{\frac{\pi_\theta\left(\vz_{i, t} \mid \vx, \vz_{i,<t}\right)}{\pi_{\theta_{\text{old}}}\left(\vz_{i, t} \mid \vx, \vz_{i,<t}\right)}\bigg\} A_i + \sum_{t'=1}^{\left|\vy^\star\right|}\text{Clip}\bigg\{\frac{\pi_\theta\left(\vy^\star_{t} \mid \vx, \vz_i\right)}{\pi_{\theta_{\text{old}}}\left(\vy^\star_{t} \mid \vx, \vz_i\right)}\bigg\} R_i\Bigg],
\end{align*}
where $\pi_{\theta_{\text{old}}}$ is the sampling policy, $A_i = \pi_{\theta_{\text{old}}}(\vy^\star |\vx, \vz_i) - \frac{1}{G-1}\sum_{j\neq i} \pi_{\theta_{\text{old}}}(\vy^\star |\vx, \vz_j)$, $R_i = \pi_{\theta_{\text{old}}}(\vy^\star |\vx, \vz_i)$, and $\text{Clip}\{\cdot\}$ denotes the PPO clipping operation.

%% file: sections/appendix/A20-dataset.tex
\section{Dataset Details}

\begin{figure*}[h!]
\centering
\includegraphics[width=0.90\textwidth]{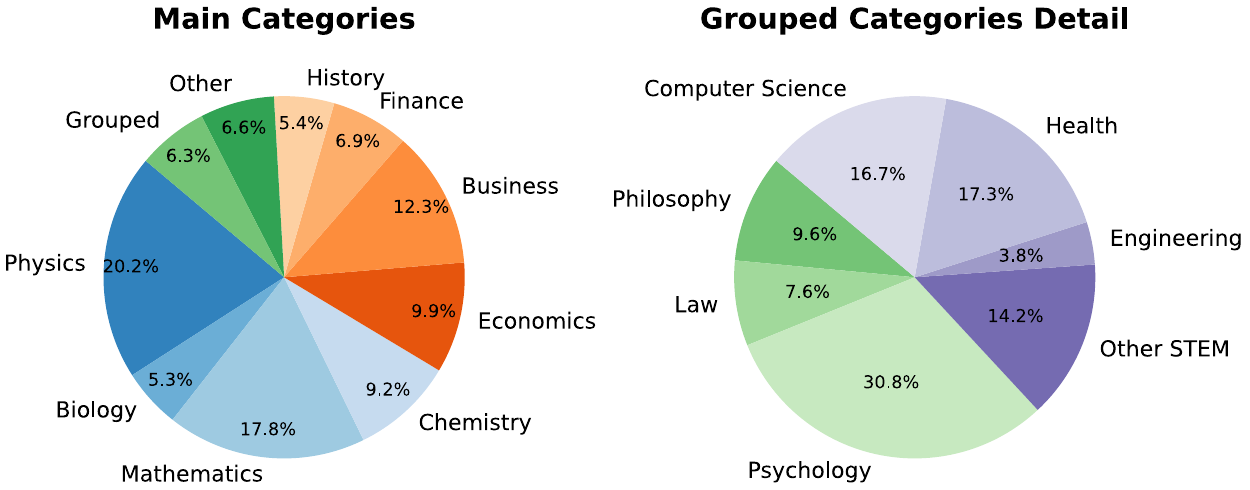}
  \caption{Category distributions in WebData. A breakdown of the ``grouped'' category (right) shows detailed distribution of various domains with fewer data samples.}{\label{fig:webdata_category_distribution}}
\end{figure*}

The category distribution in WebData (our training data) is visualized in \Cref{fig:webdata_category_distribution}.

%% file: sections/appendix/A30-extended-results.tex
\section{Extended Empirical Results}
\label{app:more_results}

In \Cref{tab:mmlu_pro,tab:superGPQA}, we provide detailed benchmark results for MMLU-Pro and SuperGPQA using domain name abbreviations. The full nomenclature is as follows:

\noindent\textbf{MMLU-Pro} (\Cref{tab:mmlu_pro}): \\
\indent CS (Computer Science), Math (Mathematics), Chem (Chemistry), Eng (Engineering), \\
\indent Law (Law), Bio (Biology), Health (Health), Phys (Physics), Bus (Business), \\
\indent Phil (Philosophy), Econ (Economics), Other (Other), Psy (Psychology), Hist (History).

\noindent\textbf{SuperGPQA} (\Cref{tab:superGPQA}): \\
\indent Eng. (Engineering), Med. (Medicine), Sci. (Science), Phil. (Philosophy), \\
\indent M.S. (Military Science), Econ. (Economics), Mgmt. (Management), Socio. (Sociology), \\
\indent L/A (Literature and Arts), Hist. (History), Agron. (Agronomy), Law (Law), Edu (Education).

In \Cref{tab:math}, we present an accuracy comparison across six math evaluation benchmarks and GPQA. Models trained with \method{} demonstrate consistent and significant improvements over the base models, further validating the effectiveness of our approach.

\begin{table}[h]
    \fontsize{7.5}{9}\selectfont
    \tabcolsep 2.0pt
    \renewcommand{\arraystretch}{1.0}
    \caption{Accuracy comparison on math evaluation suite and GPQA.}
    \label{tab:math}
    \centering
    \input{tables/math}
    \vspace{-.3cm}
\end{table}

%% file: tables/math.tex
\begin{tabular}{l|cc|cc|cc|cc|cc|cc|cc}
\toprule
\multirow{2}{*}{Method}  & \multicolumn{2}{c|}{AIME24}  & \multicolumn{2}{c|}{AMC}
& \multicolumn{2}{c|}{GSM8K}
& \multicolumn{2}{c|}{MATH-500}
& \multicolumn{2}{c|}{Minerva}
& \multicolumn{2}{c|}{Olympiad}
& \multicolumn{2}{c}{GPQA}
\\
\cmidrule{2-3} 
\cmidrule{4-5} 
\cmidrule{6-7}  
\cmidrule{8-9} 
\cmidrule{10-11} 
\cmidrule{12-13} 
\cmidrule{14-15}
& \multicolumn{1}{c}{Acc.} & \multicolumn{1}{c|}{Len.}
& \multicolumn{1}{c}{Acc.} & \multicolumn{1}{c|}{Len.}
& \multicolumn{1}{c}{Acc.} & \multicolumn{1}{c|}{Len.}
& \multicolumn{1}{c}{Acc.} & \multicolumn{1}{c|}{Len.}
& \multicolumn{1}{c}{Acc.} & \multicolumn{1}{c|}{Len.}
& \multicolumn{1}{c}{Acc.} & \multicolumn{1}{c|}{Len.}
& \multicolumn{1}{c}{Acc.} & \multicolumn{1}{c}{Len.}
\\
\midrule
Qwen3-1.7B-Base 	&1.7	&1287	&40.0	&1001	&71.4	&343	&58.8	&1094	&19.5	&1071	&23.3	&1737	&17.7	&1355		 \\
Qwen3-1.7B (w/o thinking)	&10.9	&2552	&40.0	&2291	&83.3	&312	&72.8	&1021	&27.9	&776	&39.1	&1970	&19.2	&1885		 \\
Qwen3-1.7B (w/ thinking)	&20.9	&7855	&57.5	&5973	&89.0	&2220	&77.4	&4525	&36.8	&5606	&40.9		&6497 &17.7	&7104\\
Qwen3-1.7B-Base-Verifier	&8.4	&2317	&42.5	&1712	&81.7	&405	&66.2	&1057	&30.5	&1357	&29.8	&1885			&36.4	&1317	\\	
\rowcolor{table-blue!66}Qwen3-1.7B-Base-VeriFree	&10.7	&1783	&37.5	&1522	&76.2	&447	&63.8	&972	&21.0	&867	&29.8	&1484	&30.3	&1147	\\												

\midrule

Qwen3-4B-Base	&5.2	&1312	&50.0	&1001	&73.1	&393	  &73.4	&724	&29.4	&1027	&39.6	&1202			&24.7	&1275		\\					
Qwen3-4B (w/o thinking)	&20.4	&3146	&67.5	&2047	&92.1	&308	&82.2	&1143	&41.2	&822	&49.5	&2593		&29.8	&2127	\\					
Qwen3-4B (w/ thinking)	&33.0  &7750	&62.5	&6123	&92.9	&2261	&84.4	&4370	&41.5	&5352	&47.3	&6432	&31.8	&6504	\\	

Qwen3-4B-Base-Verifier	&15.6	&2407	&57.5	&1566	&72.5	&377	&73.8	&1023	&24.6	&1088	&45.9	&1576		&44.4	&1266	\\

\rowcolor{table-blue!66}Qwen3-4B-Base-VeriFree	&16.9	&2706	&65.0	&1904	&87.5	&682	&74.8	&1269	&25.4	&1444	&44.9	&1899 &42.4	& 1619\\

\midrule

Qwen2.5-7B	&2.2	&869	&32.5	&922	&84.6	&260	&63.2	&582	&26.8	&819	&30.2	&963 &24.2	&595	\\
Qwen-2.5-7B-SimpleRL-Zoo	&15.5&	1285	&57.5	&1097	&92.1	&331	&78.6	&690	&36.4	&795	&43.4	&1083 &23.7	&990	\\
Qwen2.5-Math-7B-Oat-Zero	&28.3	&1115	&65.0	&846	&90.8&	386	&79.0	&652	&33.1	&655	&43.0	&860 &24.7	&722	\\
Qwen2.5-7B-Instruct	&11.2	&993	&52.5	&986	&91.7&	318	&78.2	&649	&37.9	&690&	39.9&	1123 &31.8	&643	\\

General-Reasoner-7B	&13.1	&1363	&52.5	&1083	&81.7	&408	&74.6	&858	&23.5	&968	&39.3	&1303	&34.8	&1199 \\
Qwen3-8B-Base	&6.5	&1213&	65.0	&917	&91.7	&304	&77.0	&671	&38.2	&884	&41.3	&1189	&38.9	&887 \\
Qwen3-8B (w/o thinking)	&24.6	&2897	&62.5	&1919	&93.3	&293	&82.2	&1087	&36.4	&751	&49.5	&2422		&27.8	&2204		\\					
Qwen3-8B (w/ thinking)	&31.0	&7789	&62.5&	6201&	95.3&	2203	&83.0	&4414&	43.8&	5410&	45.2&	6554 &37.4	&6681 \\

Qwen3-8B-Base-Verifier &17.8	&1954	&57.5	&1265	&93.4	&307	&76.2	&784	&39.0	&698	&41.8	&1201	& 43.9	&733					\\										
\rowcolor{table-blue!66}Qwen3-8B-Base-VeriFree     & 25.2 & 2461  & 67.5 & 1542 & 91.6&	375	&80.8	&1024	&30.5	&902	&50.2	&1778  &44.4	&1080\\
\bottomrule
\end{tabular}

%% file: sections/appendix/A40-case-study.tex
\section{Case Study}
In this section, we provide some examples of reasoning traces produced by the model trained by our method on training questions. And we also provide an example with both our model's and base model's responses on the test dataset.

\begin{AIbox}{Example question in training dataset (category: Chemistry)}
\textbf{Prompt}:\\
<|im\_start|>user\\
Label or a placard colored black and white with a skull and crossbones falls under which class\\
Please reason step by step, and put your final answer within <answer> {\textbackslash}boxed{} </answer>.<|im\_end|>\\
<|im\_start|>assistant 
\\
\\
\\
\textbf{Our Model's Response}:\\
A label or placard colored black and white with a skull and crossbones indicates the class of hazardous materials known as **Class 6**. Class 6 is specifically divided into two divisions:
\\\\
- **Division 6.1**: Toxic substances. These are materials that can cause injury or death if ingested or inhaled.\\
- **Division 6.2**: Infectious substances. These are materials that contain pathogens and can cause disease in humans or animals.
\\\\
The skull and crossbones symbol is universally recognized as representing toxicity, which corresponds to **Division 6.1** of Class 6 hazardous materials. <answer> {\textbackslash}boxed\{Class 6\} </answer>
\\
\\
\textbf{Reference Answer}:\\
Class 6
\end{AIbox}

\begin{AIbox}{Example question in training dataset (category: Physics)}
\textbf{Prompt}:\\
<|im\_start|>user

For what type of object is the period-luminosity relation used for determining distances?                                                                                                                        
Please reason step by step, and put your final answer within <answer> {\textbackslash}boxed\{\} </answer>.<|im\_end|>

<|im\_start|>assistant 
\\
\\
\\
\textbf{Our Model's Response}:\\
The period-luminosity relation is a fundamental tool in astronomy used to determine distances, particularly for a specific type of star known as Cepheid variables.
\\
\\
Cepheid variables are pulsating stars that exhibit a regular pattern of expansion and contraction. This pulsation causes them to vary in luminosity, and the key insight about these stars is that the period of 
their luminosity variation is directly related to their intrinsic luminosity (absolute magnitude). In other words, the longer the period of pulsation, the more intrinsically luminous the star is.
\\
\\
This period-luminosity relation was first discovered by Henrietta Leavitt in 1908 by studying Cepheid variables in the Large Magellanic Cloud. Because Cepheids are so bright intrinsically, they can be used as "standard candles" to measure distances within our galaxy and to nearby galaxies.
\\
\\
Here's how the period-luminosity relation is used to determine distances:
\\
\\
1. **Observe the Period**: First, astronomers measure the period of the star's brightness variation, which is the time it takes for the star to go from brightest to dimmest and back to brightest.
\\
\\
2. **Determine Intrinsic Luminosity**: Using the period-luminosity relation, they can then determine the star's intrinsic luminosity. This relation is usually given in a form where a graph or a formula relates period to absolute magnitude (a measure of intrinsic brightness).
\\
\\
3. **Compare to Apparent Luminosity**: The apparent magnitude (how bright the star appears from Earth) can be measured directly.
\\
\\
4. **Calculate Distance**: Using the inverse square law for light, which relates the intrinsic luminosity, apparent brightness, and distance, astronomers can then calculate the distance to the Cepheid variable.
\\
\\
This method is crucial for measuring distances in astronomy because it provides a way to gauge how far away these stars are with a high degree of accuracy. The distances determined using Cepheid variables are themselves used to calibrate other distance measurement techniques, such as those involving supernovae.
\\
\\
Given these details, the type of object for which the period-luminosity relation is used for determining distances is Cepheid variables. <answer> {\textbackslash}boxed\{Cepheid variables\} </answer>
\\
\\
\textbf{Reference Answer}:\\
cepheid variable stars
\end{AIbox}

\begin{AIbox}{Example question in MMLU-Pro benchmark (category: Psychology)}
\textbf{Prompt} (Here we ignore the prompt template and only show the question for brevity):\\
During the second stage of Kohlberg’s preconventional level of moral development, children obey rules because:\\
Options are:\\
A. they are taught that rules are unbreakable.\\
B. they enjoy following the rules.\\
C. they are rewarded for following rules.\\
D. they feel they have a personal duty to uphold rules and laws.\\
E. doing so helps them satisfy their personal needs.\\
F. they understand the concept of fairness and justice.\\
G. doing so helps them avoid punishment.\\
H. """everyone else is doing it."""\\
I. they believe in the inherent goodness of rules.\\
J. they fear societal disapproval.\\
\\
\textbf{Our Model's Response}:\\
In Kohlberg’s theory of moral development, the preconventional level is the first stage of moral reasoning, and it is primarily focused on self-interest. This level is divided into two stages:
\\\\
1. **Stage 1: Obedience and Punishment Orientation**\\
\indent ~~~~In this stage, children obey rules to avoid punishment. Their reasoning is fear-based, and they view rules as set by authority figures. The focus is on the consequences of following or breaking rules, speci
fically the avoidance of punishment.
\\
\\2. **Stage 2: Individualism and Exchange**  \\
\indent ~~~~In this stage, which is the second stage of the preconventional level, children recognize that different people have different needs and interests. They obey rules only when it is in their own interest or when they can benefit from doing so. This stage is often described as the "what's in it for me?" stage. Children at this level understand the concept of exchange and may follow rules if it leads to a reciprocal benefit.
\\\\
Now, let's evaluate the options:
\\\\
(A) they are taught that rules are unbreakable.\\                 
\indent ~~~~ - This does not align with either stage of the preconventional level.
\\\\
(B) they enjoy following the rules.\\
\indent ~~~~ - Enjoyment of rules is more characteristic of conventional levels of moral development.
\\\\
(C) they are rewarded for following rules.\\
\indent ~~~~ - While this could be somewhat related to Stage 2, the primary focus in Stage 2 is on personal benefit rather than external rewards.
\\\\
(D) they feel they have a personal duty to uphold rules and laws.  \\
\indent ~~~~ - This is more aligned with the conventional level, specifically the intermediate stage.
\\\\
(E) doing so helps them satisfy their personal needs.                      
\\
\indent ~~~~ - This could be seen as related to Stage 2, but it is not the most precise description.  
\\\\
(F) they understand the concept of fairness and justice.\\  
\indent ~~~~ - Understanding fairness and justice is more associated with the conventional and post-conventional levels.
\\\\
(G) they avoid punishment.  \\
\indent ~~~~ - This describes Stage 1 of the preconventional level, not Stage 2.
\end{AIbox}

\begin{AIbox}{Cont.}
(H) everyone else is doing it.\\
\indent ~~~~ - This is more related to conformity and is not a key characteristic of Kohlberg’s stages.
\\\\
(I) they believe in the inherent goodness of rules.\\
\indent ~~~~ - This is more characteristic of the post-conventional level.
\\\\
(J) they fear societal disapproval.\\
\indent ~~~~ - Fear of societal disapproval is more related to the conventional level.
\\\\
The best fit for Stage 2 of the preconventional level is (E) doing so helps them satisfy their personal needs, as it reflects the idea of mutual benefit and personal interest. However, (G) they avoid punishment describes Stage 1 more accurately. Given the typical interpretations of Kohlberg's stages, (E) is the more appropriate choice for Stage 2.
\\\\
But let's double-check: According to Kohlberg, in Stage 2, children's moral reasoning is based on the notion of exchange and the satisfaction of self-interest. Therefore, (E) is the most accurate option.
\\\\
<answer> {\textbackslash}boxed\{E\} </answer>
\\\\
\textbf{Base Model's Response}:\\
During the second stage of Kohlberg’s preconventional level of moral development, children obey rules because they understand the concept of fairness and justice. This stage is characterized by a focus on the consequences of actions and the desire to avoid punishment or gain rewards. Children at this stage begin to recognize that rules are not absolute and can be negotiated based on the situation and the needs of the individual. They start to consider the perspectives of others and the fairness of the rules, which is a significant step towards more advanced moral reasoning.
\\\\
Therefore, the correct answer is:
\\\\
{\textbackslash}boxed\{F\}
\\\\
\textbf{Reference Answer}:\\
E
\end{AIbox}

%% file: main.bbl
\begin{thebibliography}{52}
\providecommand{\natexlab}[1]{#1}
\providecommand{\url}[1]{\texttt{#1}}
\expandafter\ifx\csname urlstyle\endcsname\relax
  \providecommand{\doi}[1]{doi: #1}\else
  \providecommand{\doi}{doi: \begingroup \urlstyle{rm}\Url}\fi

\bibitem[Ahmadian et~al.(2024)Ahmadian, Cremer, Gall{\'e}, Fadaee, Kreutzer, Pietquin, {\"U}st{\"u}n, and Hooker]{ahmadian2024back}
Arash Ahmadian, Chris Cremer, Matthias Gall{\'e}, Marzieh Fadaee, Julia Kreutzer, Olivier Pietquin, Ahmet {\"U}st{\"u}n, and Sara Hooker.
\newblock Back to basics: Revisiting reinforce style optimization for learning from human feedback in llms.
\newblock \emph{arXiv preprint arXiv:2402.14740}, 2024.

\bibitem[Casella and Robert(1996)]{raolackwellisation}
George Casella and Christian~P. Robert.
\newblock Rao-blackwellisation of sampling schemes.
\newblock \emph{Biometrika}, 83\penalty0 (1):\penalty0 81--94, 1996.

\bibitem[Chen et~al.(2025)Chen, Liu, Du, Pang, Liu, Sinha, Varakantham, and Lin]{chen2025bootstrapping}
Changyu Chen, Zichen Liu, Chao Du, Tianyu Pang, Qian Liu, Arunesh Sinha, Pradeep Varakantham, and Min Lin.
\newblock Bootstrapping language models with dpo implicit rewards.
\newblock In \emph{International Conference on Learning Representations (ICLR)}, 2025.

\bibitem[Chen et~al.(2024)Chen, Feng, Liu, Yao, Prabhakar, Heinecke, Ho, Mui, Savarese, Xiong, et~al.]{chen2024language}
Haolin Chen, Yihao Feng, Zuxin Liu, Weiran Yao, Akshara Prabhakar, Shelby Heinecke, Ricky Ho, Phil Mui, Silvio Savarese, Caiming Xiong, et~al.
\newblock Language models are hidden reasoners: Unlocking latent reasoning capabilities via self-rewarding.
\newblock \emph{arXiv preprint arXiv:2411.04282}, 2024.

\bibitem[Cobbe et~al.(2021)Cobbe, Kosaraju, Bavarian, Chen, Jun, Kaiser, Plappert, Tworek, Hilton, Nakano, et~al.]{cobbe2021training}
Karl Cobbe, Vineet Kosaraju, Mohammad Bavarian, Mark Chen, Heewoo Jun, Lukasz Kaiser, Matthias Plappert, Jerry Tworek, Jacob Hilton, Reiichiro Nakano, et~al.
\newblock Training verifiers to solve math word problems.
\newblock \emph{arXiv preprint arXiv:2110.14168}, 2021.

\bibitem[Du et~al.(2025)Du, Yao, Ma, Wang, Zheng, Zhu, Liu, Liang, Jin, Wei, et~al.]{du2025supergpqa}
Xinrun Du, Yifan Yao, Kaijing Ma, Bingli Wang, Tianyu Zheng, King Zhu, Minghao Liu, Yiming Liang, Xiaolong Jin, Zhenlin Wei, et~al.
\newblock Supergpqa: Scaling llm evaluation across 285 graduate disciplines.
\newblock \emph{arXiv preprint arXiv:2502.14739}, 2025.

\bibitem[Gao et~al.(2023)Gao, Schulman, and Hilton]{gao2023scaling}
Leo Gao, John Schulman, and Jacob Hilton.
\newblock Scaling laws for reward model overoptimization.
\newblock In \emph{International Conference on Machine Learning}, pages 10835--10866. PMLR, 2023.

\bibitem[Gehring et~al.(2024)Gehring, Zheng, Copet, Mella, Carbonneaux, Cohen, and Synnaeve]{gehring2024rlef}
Jonas Gehring, Kunhao Zheng, Jade Copet, Vegard Mella, Quentin Carbonneaux, Taco Cohen, and Gabriel Synnaeve.
\newblock Rlef: Grounding code llms in execution feedback with reinforcement learning.
\newblock \emph{arXiv preprint arXiv:2410.02089}, 2024.

\bibitem[Grattafiori et~al.(2024)Grattafiori, Dubey, Jauhri, Pandey, Kadian, Al-Dahle, Letman, Mathur, Schelten, Vaughan, et~al.]{grattafiori2024llama}
Aaron Grattafiori, Abhimanyu Dubey, Abhinav Jauhri, Abhinav Pandey, Abhishek Kadian, Ahmad Al-Dahle, Aiesha Letman, Akhil Mathur, Alan Schelten, Alex Vaughan, et~al.
\newblock The llama 3 herd of models.
\newblock \emph{arXiv preprint arXiv:2407.21783}, 2024.

\bibitem[Guo et~al.(2025)Guo, Yang, Zhang, Song, Zhang, Xu, Zhu, Ma, Wang, Bi, et~al.]{guo2025deepseek}
Daya Guo, Dejian Yang, Haowei Zhang, Junxiao Song, Ruoyu Zhang, Runxin Xu, Qihao Zhu, Shirong Ma, Peiyi Wang, Xiao Bi, et~al.
\newblock Deepseek-r1: Incentivizing reasoning capability in llms via reinforcement learning.
\newblock \emph{arXiv preprint arXiv:2501.12948}, 2025.

\bibitem[He et~al.(2024)He, Luo, Bai, Hu, Thai, Shen, Hu, Han, Huang, Zhang, Liu, Qi, Liu, and Sun]{he2024olympiadbench}
Chaoqun He, Renjie Luo, Yuzhuo Bai, Shengding Hu, Zhen~Leng Thai, Junhao Shen, Jinyi Hu, Xu~Han, Yujie Huang, Yuxiang Zhang, Jie Liu, Lei Qi, Zhiyuan Liu, and Maosong Sun.
\newblock Olympiadbench: A challenging benchmark for promoting agi with olympiad-level bilingual multimodal scientific problems, 2024.

\bibitem[Hendrycks et~al.(2021{\natexlab{a}})Hendrycks, Burns, Kadavath, Arora, Basart, Tang, Song, and Steinhardt]{hendrycks2021measuring}
Dan Hendrycks, Collin Burns, Saurav Kadavath, Akul Arora, Steven Basart, Eric Tang, Dawn Song, and Jacob Steinhardt.
\newblock Measuring mathematical problem solving with the math dataset.
\newblock \emph{arXiv preprint arXiv:2103.03874}, 2021{\natexlab{a}}.

\bibitem[Hendrycks et~al.(2021{\natexlab{b}})Hendrycks, Burns, Kadavath, Arora, Basart, Tang, Song, and Steinhardt]{hendrycksmath2021}
Dan Hendrycks, Collin Burns, Saurav Kadavath, Akul Arora, Steven Basart, Eric Tang, Dawn Song, and Jacob Steinhardt.
\newblock Measuring mathematical problem solving with the math dataset.
\newblock \emph{NeurIPS}, 2021{\natexlab{b}}.

\bibitem[Hu et~al.(2025)Hu, Zhang, Han, Jiang, Zhang, and Shum]{hu2025openreasonerzeroopensourceapproach}
Jingcheng Hu, Yinmin Zhang, Qi~Han, Daxin Jiang, Xiangyu Zhang, and Heung-Yeung Shum.
\newblock Open-reasoner-zero: An open source approach to scaling up reinforcement learning on the base model, 2025.
\newblock URL \url{https://arxiv.org/abs/2503.24290}.

\bibitem[Huang et~al.(2024)Huang, Block, Foster, Rohatgi, Zhang, Simchowitz, Ash, and Krishnamurthy]{huang2024self}
Audrey Huang, Adam Block, Dylan~J Foster, Dhruv Rohatgi, Cyril Zhang, Max Simchowitz, Jordan~T Ash, and Akshay Krishnamurthy.
\newblock Self-improvement in language models: The sharpening mechanism.
\newblock \emph{arXiv preprint arXiv:2412.01951}, 2024.

\bibitem[Kwon et~al.(2023)Kwon, Li, Zhuang, Sheng, Zheng, Yu, Gonzalez, Zhang, and Stoica]{kwon2023efficient}
Woosuk Kwon, Zhuohan Li, Siyuan Zhuang, Ying Sheng, Lianmin Zheng, Cody~Hao Yu, Joseph~E. Gonzalez, Hao Zhang, and Ion Stoica.
\newblock Efficient memory management for large language model serving with pagedattention.
\newblock In \emph{Proceedings of the ACM SIGOPS 29th Symposium on Operating Systems Principles}, 2023.

\bibitem[Lambert et~al.(2024)Lambert, Morrison, Pyatkin, Huang, Ivison, Brahman, Miranda, Liu, Dziri, Lyu, et~al.]{lambert2024tulu3}
Nathan Lambert, Jacob Morrison, Valentina Pyatkin, Shengyi Huang, Hamish Ivison, Faeze Brahman, Lester James~V Miranda, Alisa Liu, Nouha Dziri, Shane Lyu, et~al.
\newblock T$\backslash$" ulu 3: Pushing frontiers in open language model post-training.
\newblock \emph{arXiv preprint arXiv:2411.15124}, 2024.

\bibitem[Lewkowycz et~al.(2022)Lewkowycz, Andreassen, Dohan, Dyer, Michalewski, Ramasesh, Slone, Anil, Schlag, Gutman-Solo, et~al.]{lewkowycz2022solving}
Aitor Lewkowycz, Anders Andreassen, David Dohan, Ethan Dyer, Henryk Michalewski, Vinay Ramasesh, Ambrose Slone, Cem Anil, Imanol Schlag, Theo Gutman-Solo, et~al.
\newblock Solving quantitative reasoning problems with language models.
\newblock \emph{Advances in Neural Information Processing Systems}, 35:\penalty0 3843--3857, 2022.

\bibitem[Li et~al.(2024)Li, Beeching, Tunstall, Lipkin, Soletskyi, Huang, Rasul, Yu, Jiang, Shen, et~al.]{li2024numinamath}
Jia Li, Edward Beeching, Lewis Tunstall, Ben Lipkin, Roman Soletskyi, Shengyi Huang, Kashif Rasul, Longhui Yu, Albert~Q Jiang, Ziju Shen, et~al.
\newblock Numinamath: The largest public dataset in ai4maths with 860k pairs of competition math problems and solutions.
\newblock \emph{Hugging Face repository}, 13:\penalty0 9, 2024.

\bibitem[Lightman et~al.(2024)Lightman, Kosaraju, Burda, Edwards, Baker, Lee, Leike, Schulman, Sutskever, and Cobbe]{lightman2024lets}
Hunter Lightman, Vineet Kosaraju, Yuri Burda, Harrison Edwards, Bowen Baker, Teddy Lee, Jan Leike, John Schulman, Ilya Sutskever, and Karl Cobbe.
\newblock Let's verify step by step.
\newblock In \emph{The Twelfth International Conference on Learning Representations}, 2024.
\newblock URL \url{https://openreview.net/forum?id=v8L0pN6EOi}.

\bibitem[Liu et~al.(2025{\natexlab{a}})Liu, Zhang, Qin, Ossowski, Gu, Jin, Kiblawi, Preston, Wei, Vozila, Naumann, and Poon]{liu2025xreasonergeneralizablereasoningmodalities}
Qianchu Liu, Sheng Zhang, Guanghui Qin, Timothy Ossowski, Yu~Gu, Ying Jin, Sid Kiblawi, Sam Preston, Mu~Wei, Paul Vozila, Tristan Naumann, and Hoifung Poon.
\newblock X-reasoner: Towards generalizable reasoning across modalities and domains, 2025{\natexlab{a}}.
\newblock URL \url{https://arxiv.org/abs/2505.03981}.

\bibitem[Liu et~al.(2024)Liu, Chen, Du, Lee, and Lin]{liu2024oat}
Zichen Liu, Changyu Chen, Chao Du, Wee~Sun Lee, and Min Lin.
\newblock Oat: A research-friendly framework for llm online alignment.
\newblock \url{https://github.com/sail-sg/oat}, 2024.

\bibitem[Liu et~al.(2025{\natexlab{b}})Liu, Chen, Li, Pang, Du, and Lin]{liu2025there}
Zichen Liu, Changyu Chen, Wenjun Li, Tianyu Pang, Chao Du, and Min Lin.
\newblock There may not be aha moment in r1-zero-like training — a pilot study.
\newblock \url{https://oatllm.notion.site/oat-zero}, 2025{\natexlab{b}}.
\newblock Notion Blog.

\bibitem[Liu et~al.(2025{\natexlab{c}})Liu, Chen, Li, Qi, Pang, Du, Lee, and Lin]{liu2025understanding}
Zichen Liu, Changyu Chen, Wenjun Li, Penghui Qi, Tianyu Pang, Chao Du, Wee~Sun Lee, and Min Lin.
\newblock Understanding r1-zero-like training: A critical perspective.
\newblock \emph{arXiv preprint arXiv:2503.20783}, 2025{\natexlab{c}}.

\bibitem[Luo et~al.(2025{\natexlab{a}})Luo, Tan, Huang, Patel, Ariyak, Wu, Shi, Xin, Cai, Weber, Zhang, Li, Popa, and Stoica]{deepcoder2025}
Michael Luo, Sijun Tan, Roy Huang, Ameen Patel, Alpay Ariyak, Qingyang Wu, Xiaoxiang Shi, Rachel Xin, Colin Cai, Maurice Weber, Ce~Zhang, Li~Erran Li, Raluca~Ada Popa, and Ion Stoica.
\newblock Deepcoder: A fully open-source 14b coder at o3-mini level, 2025{\natexlab{a}}.
\newblock Notion Blog.

\bibitem[Luo et~al.(2025{\natexlab{b}})Luo, Tan, Wong, Shi, Tang, Roongta, Cai, Luo, Li, Popa, and Stoica]{deepscaler2025}
Michael Luo, Sijun Tan, Justin Wong, Xiaoxiang Shi, William~Y. Tang, Manan Roongta, Colin Cai, Jeffrey Luo, Li~Erran Li, Raluca~Ada Popa, and Ion Stoica.
\newblock Deepscaler: Surpassing o1-preview with a 1.5b model by scaling rl, 2025{\natexlab{b}}.
\newblock Notion Blog.

\bibitem[Ma et~al.(2025)Ma, Liu, Jiang, Zhang, Ma, and Chen]{generalreasoner}
Xueguang Ma, Qian Liu, Dongfu Jiang, Ge~Zhang, Zejun Ma, and Wenhu Chen.
\newblock General-reasoner: Advancing llm reasoning across all domains.
\newblock \emph{arXiv preprint arXiv:2505.14652}, 2025.

\bibitem[Ng et~al.(1999)Ng, Harada, and Russell]{rewardshaping2}
A.~Ng, Daishi Harada, and Stuart~J. Russell.
\newblock Policy invariance under reward transformations: Theory and application to reward shaping.
\newblock In \emph{International Conference on Machine Learning}, 1999.
\newblock URL \url{https://api.semanticscholar.org/CorpusID:5730166}.

\bibitem[{OpenAI}(2024)]{o1}
{OpenAI}.
\newblock Learning to reason with llms, 2024.
\newblock URL \url{https://openai.com/index/learning-to-reason-with-llms/}.

\bibitem[Ouyang et~al.(2022)Ouyang, Wu, Jiang, Almeida, Wainwright, Mishkin, Zhang, Agarwal, Slama, Ray, et~al.]{ouyang2022training}
Long Ouyang, Jeffrey Wu, Xu~Jiang, Diogo Almeida, Carroll Wainwright, Pamela Mishkin, Chong Zhang, Sandhini Agarwal, Katarina Slama, Alex Ray, et~al.
\newblock Training language models to follow instructions with human feedback.
\newblock \emph{Advances in neural information processing systems}, 35:\penalty0 27730--27744, 2022.

\bibitem[Pan et~al.(2025)Pan, Zhang, Wang, Yuan, Peng, and Suhr]{tinyzero}
Jiayi Pan, Junjie Zhang, Xingyao Wang, Lifan Yuan, Hao Peng, and Alane Suhr.
\newblock Tinyzero.
\newblock https://github.com/Jiayi-Pan/TinyZero, 2025.
\newblock Accessed: 2025-01-24.

\bibitem[Phan et~al.(2023)Phan, Hoffman, Dohan, Douglas, Le, Parisi, Sountsov, Sutton, Vikram, and A~Saurous]{phan2023training}
Du~Phan, Matthew~Douglas Hoffman, David Dohan, Sholto Douglas, Tuan~Anh Le, Aaron Parisi, Pavel Sountsov, Charles Sutton, Sharad Vikram, and Rif A~Saurous.
\newblock Training chain-of-thought via latent-variable inference.
\newblock \emph{Advances in Neural Information Processing Systems}, 36:\penalty0 72819--72841, 2023.

\bibitem[Rafailov et~al.(2023)Rafailov, Sharma, Mitchell, Manning, Ermon, and Finn]{rafailov2023direct}
Rafael Rafailov, Archit Sharma, Eric Mitchell, Christopher~D Manning, Stefano Ermon, and Chelsea Finn.
\newblock Direct preference optimization: Your language model is secretly a reward model.
\newblock \emph{Advances in Neural Information Processing Systems}, 36:\penalty0 53728--53741, 2023.

\bibitem[Randlov and Alstrøm(1998)]{rewardshaping}
Jette Randlov and Preben Alstrøm.
\newblock Learning to drive a bicycle using reinforcement learning and shaping.
\newblock pages 463--471, 01 1998.

\bibitem[Rein et~al.(2024)Rein, Hou, Stickland, Petty, Pang, Dirani, Michael, and Bowman]{rein2024gpqa}
David Rein, Betty~Li Hou, Asa~Cooper Stickland, Jackson Petty, Richard~Yuanzhe Pang, Julien Dirani, Julian Michael, and Samuel~R Bowman.
\newblock Gpqa: A graduate-level google-proof q\&a benchmark.
\newblock In \emph{First Conference on Language Modeling}, 2024.

\bibitem[Schulman et~al.(2017)Schulman, Wolski, Dhariwal, Radford, and Klimov]{schulman2017proximal}
John Schulman, Filip Wolski, Prafulla Dhariwal, Alec Radford, and Oleg Klimov.
\newblock Proximal policy optimization algorithms.
\newblock \emph{arXiv preprint arXiv:1707.06347}, 2017.

\bibitem[Shao et~al.(2024)Shao, Wang, Zhu, Xu, Song, Bi, Zhang, Zhang, Li, Wu, et~al.]{shao2024deepseekmath}
Zhihong Shao, Peiyi Wang, Qihao Zhu, Runxin Xu, Junxiao Song, Xiao Bi, Haowei Zhang, Mingchuan Zhang, YK~Li, Y~Wu, et~al.
\newblock Deepseekmath: Pushing the limits of mathematical reasoning in open language models.
\newblock \emph{arXiv preprint arXiv:2402.03300}, 2024.

\bibitem[Su et~al.(2025)Su, Yu, Song, Li, Mi, Tu, Zhang, and Yu]{su2025expanding}
Yi~Su, Dian Yu, Linfeng Song, Juntao Li, Haitao Mi, Zhaopeng Tu, Min Zhang, and Dong Yu.
\newblock Expanding rl with verifiable rewards across diverse domains.
\newblock \emph{arXiv preprint arXiv:2503.23829}, 2025.

\bibitem[Sutton and Barto(2018)]{sutton2018rlbook}
Richard~S. Sutton and Andrew~G. Barto.
\newblock \emph{Reinforcement Learning: An Introduction}.
\newblock The MIT Press, second edition, 2018.

\bibitem[Tang et~al.(2025)Tang, Wang, and Munos]{tang2025learning}
Yunhao Tang, Sid Wang, and R{\'e}mi Munos.
\newblock Learning to chain-of-thought with jensen's evidence lower bound.
\newblock \emph{arXiv preprint arXiv:2503.19618}, 2025.

\bibitem[Team(2025)]{qwen3}
Qwen Team.
\newblock Qwen3, April 2025.
\newblock URL \url{https://qwenlm.github.io/blog/qwen3/}.

\bibitem[Wang et~al.(2024)Wang, Ma, Zhang, Ni, Chandra, Guo, Ren, Arulraj, He, Jiang, et~al.]{wang2024mmlu}
Yubo Wang, Xueguang Ma, Ge~Zhang, Yuansheng Ni, Abhranil Chandra, Shiguang Guo, Weiming Ren, Aaran Arulraj, Xuan He, Ziyan Jiang, et~al.
\newblock Mmlu-pro: A more robust and challenging multi-task language understanding benchmark.
\newblock In \emph{The Thirty-eight Conference on Neural Information Processing Systems Datasets and Benchmarks Track}, 2024.

\bibitem[Yang et~al.(2024{\natexlab{a}})Yang, Yang, Zhang, Hui, Zheng, Yu, Li, Liu, Huang, Wei, Lin, Yang, Tu, Zhang, Yang, Yang, Zhou, Lin, Dang, Lu, Bao, Yang, Yu, Li, Xue, Zhang, Zhu, Men, Lin, Li, Xia, Ren, Ren, Fan, Su, Zhang, Wan, Liu, Cui, Zhang, and Qiu]{qwen2.5}
An~Yang, Baosong Yang, Beichen Zhang, Binyuan Hui, Bo~Zheng, Bowen Yu, Chengyuan Li, Dayiheng Liu, Fei Huang, Haoran Wei, Huan Lin, Jian Yang, Jianhong Tu, Jianwei Zhang, Jianxin Yang, Jiaxi Yang, Jingren Zhou, Junyang Lin, Kai Dang, Keming Lu, Keqin Bao, Kexin Yang, Le~Yu, Mei Li, Mingfeng Xue, Pei Zhang, Qin Zhu, Rui Men, Runji Lin, Tianhao Li, Tingyu Xia, Xingzhang Ren, Xuancheng Ren, Yang Fan, Yang Su, Yichang Zhang, Yu~Wan, Yuqiong Liu, Zeyu Cui, Zhenru Zhang, and Zihan Qiu.
\newblock Qwen2.5 technical report.
\newblock \emph{arXiv preprint arXiv:2412.15115}, 2024{\natexlab{a}}.

\bibitem[Yang et~al.(2024{\natexlab{b}})Yang, Zhang, Hui, Gao, Yu, Li, Liu, Tu, Zhou, Lin, et~al.]{yang2024qwen2math}
An~Yang, Beichen Zhang, Binyuan Hui, Bofei Gao, Bowen Yu, Chengpeng Li, Dayiheng Liu, Jianhong Tu, Jingren Zhou, Junyang Lin, et~al.
\newblock Qwen2. 5-math technical report: Toward mathematical expert model via self-improvement.
\newblock \emph{arXiv preprint arXiv:2409.12122}, 2024{\natexlab{b}}.

\bibitem[Yu et~al.(2025)Yu, Zhang, Zhu, Yuan, Zuo, Yue, Fan, Liu, Liu, Liu, et~al.]{yu2025dapo}
Qiying Yu, Zheng Zhang, Ruofei Zhu, Yufeng Yuan, Xiaochen Zuo, Yu~Yue, Tiantian Fan, Gaohong Liu, Lingjun Liu, Xin Liu, et~al.
\newblock Dapo: An open-source llm reinforcement learning system at scale.
\newblock \emph{arXiv preprint arXiv:2503.14476}, 2025.

\bibitem[Yuan et~al.(2024)Yuan, Pang, Cho, Sukhbaatar, Xu, and Weston]{yuan2024selfrewarding}
Weizhe Yuan, Richard~Yuanzhe Pang, Kyunghyun Cho, Sainbayar Sukhbaatar, Jing Xu, and Jason Weston.
\newblock Self-rewarding language models.
\newblock \emph{International Conference on Machine Learning}, 2024.
\newblock \doi{10.48550/arXiv.2401.10020}.

\bibitem[Yuan et~al.(2025)Yuan, Yu, Jiang, Padthe, Li, Wang, Kulikov, Cho, Tian, Weston, et~al.]{yuan2025naturalreasoning}
Weizhe Yuan, Jane Yu, Song Jiang, Karthik Padthe, Yang Li, Dong Wang, Ilia Kulikov, Kyunghyun Cho, Yuandong Tian, Jason~E Weston, et~al.
\newblock Naturalreasoning: Reasoning in the wild with 2.8 m challenging questions.
\newblock \emph{arXiv preprint arXiv:2502.13124}, 2025.

\bibitem[Yuan et~al.(2023)Yuan, Yuan, Li, Dong, Lu, Tan, Zhou, and Zhou]{yuan2023scalingrft}
Zheng Yuan, Hongyi Yuan, Chengpeng Li, Guanting Dong, Keming Lu, Chuanqi Tan, Chang Zhou, and Jingren Zhou.
\newblock Scaling relationship on learning mathematical reasoning with large language models.
\newblock \emph{arXiv preprint arXiv:2308.01825}, 2023.

\bibitem[Yue et~al.(2024)Yue, Zheng, Zhang, and Chen]{yue2024mammoth2}
Xiang Yue, Tianyu Zheng, Ge~Zhang, and Wenhu Chen.
\newblock Mammoth2: Scaling instructions from the web.
\newblock \emph{Advances in Neural Information Processing Systems}, 37:\penalty0 90629--90660, 2024.

\bibitem[Zeng et~al.(2025)Zeng, Huang, Liu, He, Liu, Ma, and He]{zeng2025simplerl}
Weihao Zeng, Yuzhen Huang, Wei Liu, Keqing He, Qian Liu, Zejun Ma, and Junxian He.
\newblock 7b model and 8k examples: Emerging reasoning with reinforcement learning is both effective and efficient.
\newblock \url{https://hkust-nlp.notion.site/simplerl-reason}, 2025.
\newblock Notion Blog.

\bibitem[Ziegler et~al.(2019)Ziegler, Stiennon, Wu, Brown, Radford, Amodei, Christiano, and Irving]{rlhf}
Daniel~M Ziegler, Nisan Stiennon, Jeffrey Wu, Tom~B Brown, Alec Radford, Dario Amodei, Paul Christiano, and Geoffrey Irving.
\newblock Fine-tuning language models from human preferences.
\newblock \emph{arXiv preprint arXiv:1909.08593}, 2019.

\bibitem[Zuo et~al.(2025)Zuo, Zhang, Qu, Sheng, Zhu, Qi, Sun, Cui, Ding, and Zhou]{zuo2025ttrl}
Yuxin Zuo, Kaiyan Zhang, Shang Qu, Li~Sheng, Xuekai Zhu, Biqing Qi, Youbang Sun, Ganqu Cui, Ning Ding, and Bowen Zhou.
\newblock Ttrl: Test-time reinforcement learning.
\newblock \emph{arXiv preprint arXiv:2504.16084}, 2025.

\end{thebibliography}
